\documentclass{article}

\usepackage{arxiv}

\usepackage[utf8]{inputenc} 
\usepackage[T1]{fontenc}    
\usepackage[unicode=true,pagebackref=true,
colorlinks,linkcolor=blue,citecolor=red,final]{hyperref}       
\usepackage{url}            
\usepackage{booktabs}       
\usepackage{amsfonts}       
\usepackage{nicefrac}       
\usepackage{microtype}      
\usepackage{lipsum}
\usepackage{graphicx}
\usepackage[export]{adjustbox}
\usepackage[skip=-0.3pt,font=tiny]{subcaption}
\usepackage{multirow}

\usepackage{graphicx}
\usepackage[ruled, lined]{algorithm2e}
\usepackage{booktabs}   
\usepackage{multirow} 
\usepackage{subcaption}
\usepackage{amsmath}
\usepackage{amssymb}
\usepackage{float}
\usepackage[nodisplayskipstretch]{setspace}
\graphicspath{ {./images/} }

\title{Decay2Distill: Leveraging spatial perturbation and regularization for self-supervised image denoising}

\author{
 Manisha Das Chaity \\
  Rochester Institute of Technology\\
    NY, USA \\
  \texttt{manisha.kuet@gmail.com} \\
   \And
 Masud An Nur Islam Fahim \\
  Chosun University\\
  Gwangju, South Korea \\
  \texttt{mostofafahim21@gmail.com} \\
}

\begin{document}
\maketitle
\begin{abstract}
Unpaired image denoising has achieved promising development over the last few years. Regardless of the performance, methods tend to heavily rely on underlying noise properties or any assumption which is not always practical. Alternatively, if we can ground the problem from a structural perspective rather than noise statistics, we can achieve a more robust solution. with such motivation, we propose a self-supervised denoising scheme that is unpaired and relies on spatial degradation followed by a regularized refinement. Our method shows considerable improvement over previous methods and exhibited consistent performance over different data domains.  
\end{abstract}


\section{Introduction}
Image denoising is an ever-going issue that confirms its presence in most vision problems. Anyhow, recovery of noisy observation has attracted considerable interest from the researcher and consequently, we have observed a bulk of studies, mostly dominated by the supervised approaches. Since the existence of in-domain clear observation is mostly impractical/costly, the current trend in signal recovery is dominated by self-supervised approaches. Typically we minimize ${\parallel f(x)-y\parallel}_2$ in a supervised setup, where y is the available ground truth image, f is the given estimator and x is the noisy observation. In the self-supervised setup, we minimize ${\parallel f(x)-x\parallel}_2$ with a respective novel empirical risk minimization process.

The whole development roughly started right after this 'weakly/noisy' supervised study  Noise2Noise (N2N) \cite{lehtinen2018noise2noise}, where authors obtained supervised baseline results without using a clean target. Later, Noisier2Noise (Nr2N) \cite{moran2020noisier2noise} achieves good performance just by doubling the input noise, although it's a clever modification of the N2N. Later, Noise2Self (N2S) \cite{batson2019noise2self} introduces a masking strategy to address the unpaired denoising problem, followed by similar clever alternatives 
Noise2Void (N2V) \cite{krull2019noise2void}, Noise2Same \cite{xie2020noise2same} (N2Sa), Blind2Unblind (B2Ub) \cite{wang2022blind2unblind}. Nevertheless, overall unpaired or self-paired denoising studies follow two sects in terms of algorithmic strategy  \cite{pang2021recorrupted}  : (a) signal augmentation [Nr2N, N2S,  N2V], where methods mostly focus on spatial presentation to avoid identity collapse and refined signal estimation, (b) regularization [SURE \cite{soltanayev2018training}, Noise2Fast (N2F) \cite{lequyer2021noise2fast}, Noise2Kernel (N2K) \cite{lee2020noise2kernel}]. Here, methods tweak the estimation by cleverly designed criteria with respect to the relevant hypothesis. However, the first category is limited to the principal zero mean assumption regardless of the superior performance. Regardless of the noise distribution independence, the second category suffers from limited visual fidelity or higher amounts of computation burden. We can do better and by combining sect (a) and (b) superior performance is attainable. This article proposes a novel 'push pull' strategy to address the self-paired denoising task.

In our ‘push' strategy, we perform regular augmentation-guided empirical risk minimization via ‘noise to shifted'  and ‘noise to compressed’ schemes. Such intentional degradation easily pushes the network to obtain a smoother observation. We incorporate our novel ‘pull’ strategy to obtain spatial details, which penalizes the network towards recovering fine details through complimentary shifting estimation matching. Even though our method is not independent of the zero mean noise distribution observation, our strategy can exceed the previous benchmark with the help of our 'pull' scheme and greedy yet clever 'push' scheme. More details are provided in the methodology section.
To sum up our contribution:
\begin{itemize}
 \item We propose a novel 'push-pull' training scheme that allows us to combine the best of the augmentation and regularization schemes.
 \item We show that our method benefits better optimization conditions than regular 'Noise2X’ schemes.
 \item Our strategy 'push-pull' scheme can obtain cleaner observations with details.
 \item Finally our method achieves better scores within different noise and domains. 
\end{itemize}

\section{Related Work}
\subsection{Filtering based schemes}

Noisy signal recovery is a long-studied genre. Classical signal refinement approaches like non-local mean [\cite{buades2005non}, BM3D \cite{dabov2007image}, k-SVD \cite{scetbon2021deep}] utilize a non-learning approach to estimate the smooth approximation of the noisy image. Despite their simplicity, the obtained signal often misses the fine details and decays more with stronger noise. To address the limitation with non-learning studies, paired approaches started with unprecedented success started with DnCNN \cite{zhang2017beyond}, later the improvement relayed to the [FFDNet \cite{zhang2018ffdnet}, CBDNet \cite{guo2019toward}, RIDNet \cite{zhuo2019ridnet}, Deamnet \cite{ren2021adaptive}]. The studies mentioned above utilize crafty network design, novel loss function, and ensemble modules to achieve state-of-the-art denoising results.

However, the only issue concerns the availability of the clean pair, which pushes the next phase in learning-dependent denoising studies. To address the clean pair issue, researchers focus on learning from the noisy label. However, this practice is prevalent in classification and segmentation issues with a slightly different approach. Anyhow,  N2N proposes using noisy pairs as the first alternative to the noisy-clean fashion. In practice, N2N fails to address the ground truth acquisition fully.
To address its limitation. Nr2N proposes noisy data augmentation to the input and creates a noisier-noisy pair. A similar effort is observed with the Noisy-As-Clean (NAC) \cite{elad2006image}.

To improve further, [N2S, N2V, N2Sa, B2Ub] proposes a masking strategy to tune the parameter space towards a better smoother approximation. Self2Self (S2S) \cite{quan2020self2self} uses dropout to tune the network, and N2K uses dilated convolution alternative to the S2S. Neighbor2Neighbor (Ne2Ne) \cite{huang2021neighbor2neighbor} uses spatial regularization to recover the clean signal, Recorrupted2Recorrupted (R2R)  \cite{pang2021recorrupted} is a robust extension to the Nr2N. However, all of the studies clearly broaden the path for designing a method where baseline noisy augmentation can marry the complimentary regularization scheme to achieve better results. More details on such are presented later.

\section{Methodology}
\begin{figure}[!htbp]
\begin{center}
\includegraphics[width=0.9\linewidth]{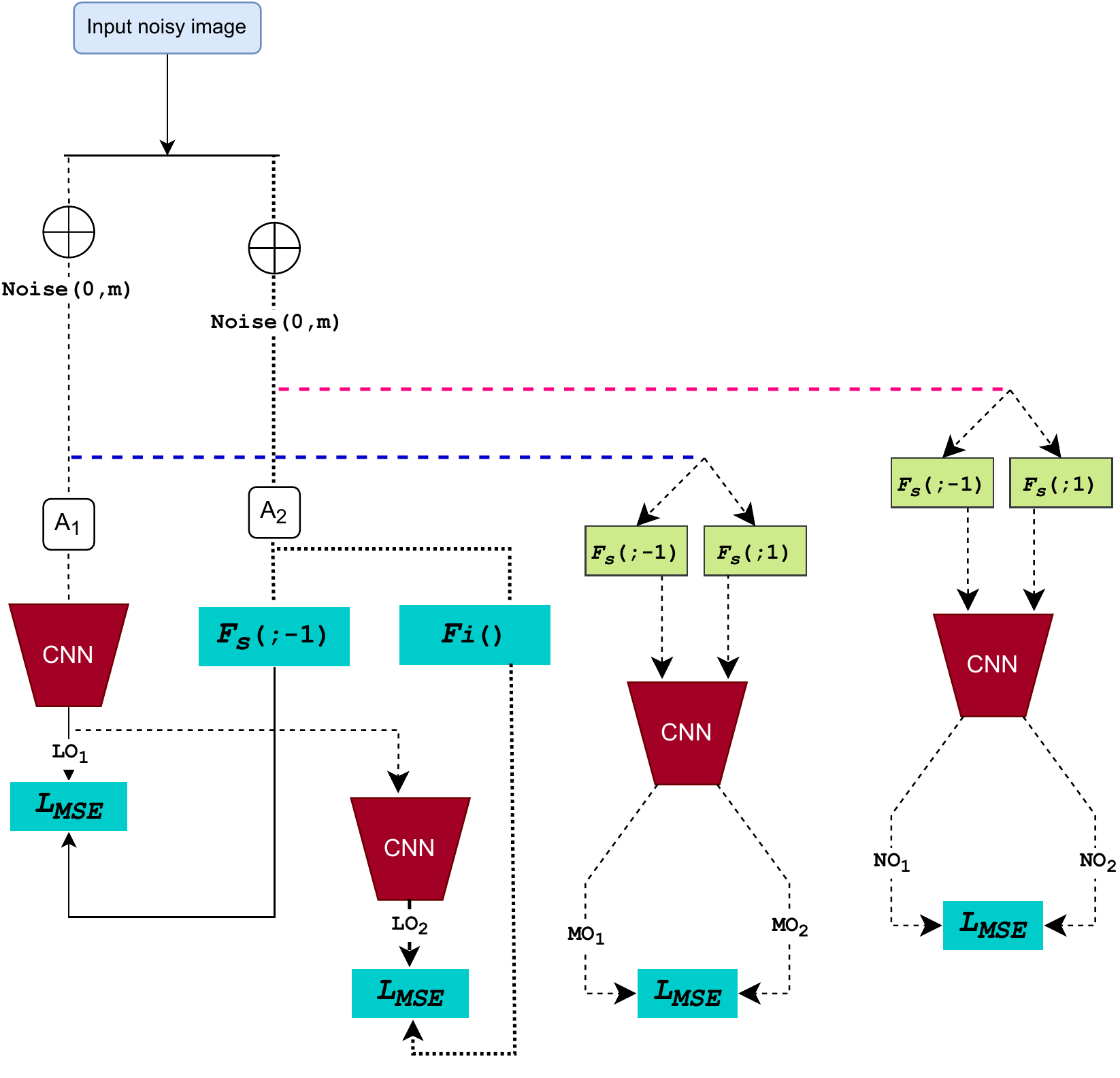}
\end{center}
   \caption{Overview of the proposed framework.}
\label{fig:1}
\end{figure}
To overcome the challenges of the previous study we assume that we don’t have access to any $J$-invariant function or multiple noisy observations or guaranteed reconstruction. To find the solution we look back at the working principle of neural networks. The neural networks are simply the refined function through averaging gradient which corresponds to the loss function. Hence, one update ensures that the next inference will satisfy the corresponding loss function on average. The simple yet powerful phenomenon encourages our proposed method. We first created two versions of the input noisy image ‘A’ adding more noise to it with the same standard deviation which is the same for synthetic or real noise. This provides benefits for zero-mean property in addition to texture preservation during inference. 

Let, our noise-added copies are $A_1$ and $A_2$. Now we will create $B_1$ by shifting a row of $A_2$ in either upward or downward direction and $B_2$ by degrading the JPEG quality of the $A_2$. Hence, our first loss function aims to map from  $A_1$ towards $B_2$ via satisfying $B_1$. More precisely, we feed $A_1$ into CNN and get intermediate output $LO_1$, and we perform MSE between $LO_1$ and $B_1$. Then we feed $LO_1$ into CNN and get $LO_2$. This time we measure MSE between $LO_2$ and $B_2$. In this way, the network learns the spatial properties of  $A_1$, $B_1$, $B_2$, and whatever lies in between $B_2$ and $B_1$. Secondly, we take $A_1$ and create $MO_1$ and $MO_2$ by random row shifting in the upward and downward direction. Similarly, we create $NO_1$ and $NO_2$ for $A_2$. As ( $MO_1$, $MO_2$) and ($NO_1$, $NO_2$) are created from the same image which implies learned representations for pairs will be the same. Hence, for our given network $f_{\theta}()$ we can write 
$f( MO_1) \approx f(MO_2)$ and $f(NO_1) \approx f(NO_2)$. We do the MSE between the pairs to introduce smoothness regularization. In summary, we update our CNN four times in a single step of smoothing, which in the end learn a smooth approximation of the clean image as the solution to the learning problem.
\subsection{Theoretical Background}
To explain our approach we first include the usual baseline for clarity. To do such we start with the Noise2Noise (N2N) \cite{lehtinen2018noise2noise}. In that paper, authors have shown that the provided algorithm is somewhat equivalent to a supervised setup. N2N requires two noisy observations $C_1$ and $C_2$ where $Var(C_1)>>Var\ (C_2)$. In other words, $C_2$ is closer to the underlying clean image compared to $C_1$. Say, x is the clean image and n is the noise. For simplicity ‘n’ is additive Gaussian noise. Hence, $C_1 = x+n_1,\ {\ n}_1\sim\mathcal{N}(0,\sigma_1^2)$ and $C_2=x+n_2,\ {\ n}_2\sim\mathcal{N}(0,\sigma_2^2)$. 
If $f_\theta( )$ is our denoising training network, then the following is the N2N setup:

\begin{equation}
\mathbb{E}_{n_1,n_2}{{\parallel f_{\theta}( C_1)- (C_2)\parallel}_2^2}=\mathbb{E}_{n_1,n_2}{{\parallel f_{\theta}(C_1)-x\parallel}_2^2}-2\mathbb{E}_{n_1,n_2}{\parallel{(n_2)}^Tf_{\theta} (C_1)\parallel}+constant
\end{equation}

As we have taken the simple yet powerful assumption that working noise is zero mean, this will lead $2\mathbb{E}_{n_1,n_2}{{\parallel{(n_2)}^T f_{\theta}(C_1)\parallel}_2^2}$\
towards zero as a result of the expectation property. This immediately turns the above equation into the desired supervised setting but is limited to a constant. 

In our work, we focused on this constant which will be discussed later. Before moving forward, we summarized the efforts of the N2N \cite{lehtinen2018noise2noise}, Nr2N \cite{moran2020noisier2noise}, NAC \cite{xu2020noisy}, and R2R \cite{pang2021recorrupted} to investigate equation (1) more closely. Although Nr2N and NAC are almost similar, the investigation is worthwhile. From equation 1 above N2N  minimizes,

\begin{equation}
\mathbb{E}_{n_1,n_2}{{\parallel{f_{\theta}}(C_1)-x\parallel}_2^2}+
\underset{constant}{\underbrace{{\parallel (C_2)-x\parallel}_2}}
\end{equation}

For Nr2N and NAC, 
$C_1^\prime=C_1+n_1^\prime$. Where, ${n_1^\prime}\sim{\mathcal{N}(0,\sigma^\prime)}$. Hence,

\begin{equation}
\mathbb{E}_{n_1,n_1^\prime}{{\parallel{f}_\theta(C_1^\prime)-C_1\parallel}_2^2}=\mathbb{E}_{n_1,n_1^\prime}{{\parallel{f}_\theta(C_1^\prime)-x\parallel}_2^2}+\underset{constant}{\underbrace{{\parallel C_1-x\parallel}_2}}    equation
\end{equation}

For R2R, $\widehat{C_1}=C_1+\widehat{n_1}$ 
where, $\widehat{n_1}\sim \mathcal{N}(0,\hat{\sigma}).$

\begin{equation}
\mathbb{E}_{n_1^\prime,\widehat{n_1}}{{\parallel{f}_\theta(\widehat{C_1})-C_1^\prime\parallel}_2^2}=\mathbb{E}_{n_1^\prime,\widehat{n_1}}{{\parallel{f}_\theta(\widehat{C_1})-x\parallel}_2^2}+\underset{constant}{\underbrace{{\parallel C_1^\prime-x\parallel}_2}}
\end{equation}

From the above method, we observe that every method aims for the limited/noisy supervised setup, which is indicated by that constant. Therefore, it is straightforward that, if the $constant \rightarrow\ 0$, the better the self-supervised methods will become and eventually becomes supervised when the constant hits zero. We take that into account and tried to reduce the effect of the constant in the first part of our loss function.

To begin, let's assume, A is the available noisy image $(A=\ x+\ n;\ n \sim\mathcal{N}(0,\sigma))$ for our experiment.  Now, the typical self-supervised setup tends to minimize ${\parallel{f}_\theta(x)-x\parallel}_2^2$\ which no doubt leads to identity. To avoid such, the common practice is designing $f_\theta(\ )$ in such a way that it blocks the identity convergence. 

Another common practice is inspired by N2V \cite{krull2019noise2void} which creates two observations of A. However, we follow the latter one and say $A_1$ and$ A_2$ are the observation that originated from A. $A_1=\ A+\ O_1;\ O_1\sim\mathcal{N}(0,\sigma)$  and $A_2=\ A+\ O_2;\ O_2\sim\mathcal{N}(0,\sigma)$. $O_1$ and $\ O_2$ are drawn with the same standard deviation and their expectations are same  $\mathbb{E}[O_1]=\mathbb{E}[O_2]=0$ but ${\parallel O_1-O_2\parallel}_2\neq0$. This will allow us to evade identity convergence phenomena.

\SetCommentSty{mycommfont}
\newlength\mylen
\newcommand\myinput[1]{%
  \settowidth\mylen{\KwIn{}}%
  \setlength\hangindent{\mylen}%
  \hspace*{\mylen}#1\\}
\SetArgSty{textnormal}
\begin{algorithm} [h!]
    \caption{Algorithm: Push and Pull}
    \SetKwFunction{isOddNumber}{isOddNumber}
    \SetKwInOut{KwIn}{Input}
    \SetKwInOut{KwOut}{Output}
\KwIn{A set of noisy image, $A_i$}
\myinput{Denoising Network, $f_\theta$}
\myinput {Stochastic Shifter, $f_S(a,d); d_k\in[1,-1]$}
\myinput {Random JPEG decay,}
\myinput { $f_J(a,p); p_k\in[0.9,1]$}
\While {\text {not converged} }{
    {Sample a noisy image $A_i\in A$}\\
    {Create $A_i^1,\ A_i^2$ by applying $\mathcal{N}(0,\sigma)$}\\
    {Generate intermediate target $B_1=f_S(A_i^2,1)$}\\
    {Generate secondary target $B_2=f_J(A_i^2,\ P_k)$}\\
    {Create shifted pair $f_S\left(A_i^1,1\right)\&\ f_S(A_i^1,-1)$}\\
    {Create shifted pair $f_S\left(A_i^2,1\right)\&\ f_S(A_i^2,-1)$}\\
    Compute push loss:\\
         { $\alpha_p={\parallel B_1-f_\theta(A_1)\parallel}_2+{\parallel B_2-f_\theta(f_\theta(A_1))\parallel}_2$}\\
         {Obtain regularized pairs:}\\
         { $\left\{MO_1,MO_2\right\}=f_\theta[f_S\left(A_i^1,1\right),f_S\left(A_i^1,-1\right)]$}\\
         { $\left\{NO_1,NO_2\right\}=f_\theta[f_S\left(A_i^2,1\right),f_S\left(A_i^2,-1\right)]$}\\
       Compute pull loss:\\
          { $\alpha_R={\parallel M O_1-MO_2\parallel}_2+{\parallel N O_1-NO_2\parallel}_2$}\\
       {Update the $f_\theta$ by minimizing:}
          {$\alpha_P+\alpha_R$}}
 \end{algorithm}
 
However, we simply don't minimize ${\parallel{f}_\theta(A_1)-A_1\parallel}_2^2$\, but we do the minimization with the help of degradation operation. First, we utilize a stochastic shifter function $f_S\left(a,d\right)$ which shifts the row of the given image ‘a' with respect to the random direction ‘d’.  $f_S\left(a,d\right)$ is stochastic because it randomly shifts 0 to 5 rows either upward or downward direction to serve the randomness. Secondly, $f_J(a,p)$ also randomly changes the JPEG quality of the given image ‘a' in accordance with the parameter ‘p’. Here, $d\ \in[1,-1]$ for either upward or downward direction and$\ p\in[0.8,1]$, where 0.8 indicates $80\%$ quality preservation, obtained empirically. 
Now, we generate two intermediate targets $B_1=f_S\left(A_2,1\right)$ and $B_2=f_j\left(A_2,P_i\right)$, then the loss function can be written as follows. Given, $D_1=f_\theta(A_1)$ and $D_2=f_\theta(D_1)$. Hence,
\begin{equation}
\alpha_s={\parallel B_1-D_1\parallel}_2+{\parallel{B}_2-D_2\parallel}_2  
\end{equation} 
If we look closely, we see that we have started from $A_1$ and aiming for $B_2$ , via $D_1$. In compact form, we aim to minimize 
${\parallel f_\theta(D_1)-B_2\parallel}_2^2$\ . If we extend it to the supervised setup then we can write,
\begin{equation}
\mathbb{E}_{\widehat{O_1},\check{O_2}}{{\parallel f_\theta{(D_1)}-x\parallel}_2^2}+{{\parallel B_2-x\parallel}_2}
\end{equation}
Now from above equation 6, our method directly guaranteed a smaller constant because ${\parallel{B_2}-x\parallel}_2<{\parallel{A_2}-x\parallel}_2$ . Even though $B_2$ is from $A_2$ and $B_2$ is the smoother version of the $A_2$,  it does provide shorter distance towards the underlying clean image $x$. Furthermore, statistically, ${Var\ [A}_2]> Var [B_2]> Var [x]$, as clean image $x$ got lesser variation. Due to this, our loss function gets a strict advantage over the previous one.

The reason we took the via towards $B_1$ is two-fold. First, minimizing ${\parallel{B_1}-D_1\parallel}_2$\ automatically gives a smoother approximation of $A_1$  because $B_2$  is constantly changing in each step of the training. Due to that, our supervised equivalent part got a shorter distance to minimize. Secondly, ${\parallel{B_2}-D_2\parallel}_2$\ allow further spatial ‘correction’, as it pushes the network to reach the slightly distorted observation of the noisy input. Applied distortion is intentional for avoiding identity mapping. Due to that, the model can also avoid typical over-smoothing if we simply do ${\parallel{f_\theta}(A_1)-B_2\parallel}_2$\ in the first place.

To sum it up, our distortion-dependent minimization allows us to change the effect of the constant and achieve smoothness in contrast to typical self-supervised baseline. As we have pushed the network to achieve a preliminary smoother, now we focus on pulling out the desirable details, which are described in the second part of this section. The second part of the cost function serves as a regularizer for the overall training process. As we have focused on smooth approximation in the first part, we need a complementary mechanism for possible spatial fidelity.

To do such, we integrate a self-regularization term that involves unmodified $A_1$ and $A_2$, in terms of information property. At first, we create two shifted versions for both $A_1$ and $A_2$. As we have mentioned earlier, $f_S\left(a,\ d\right)$ in the shifting function, where a stands for the input, d stands for the shifting direction.
Furthermore, our random shifting ranges from 0 to 5 which means sometimes the function performs no shifting at all. Anyway, we pass $f_S\left(A_1,1\right),f_S\left(A_1,-1\right),f_S\left(A_2,1\right)  \&  f_S(A_2,-1)$ through the same network and the network infers $MO_{1}, MO_2, NO_1$, and $NO_2$. Then we simply perform following:
\begin{equation}
\alpha_{R}={\parallel{MO_1}-MO_2\parallel}_2+{\parallel{NO_1}-NO_2\parallel}_2
\end{equation}
Here, we can see that $MO_1, MO_2$ and $NO_1, NO_2$ originate from the same image but they impose different representations due to the shifting. If we just do this equation (7) without shifting that will become a pure identity restoration operation. However, due to the stochastic shifting, it becomes part identity, part similarity loss. Because during training, for some images we might observe no shifting at all. Further, we do this operation on every training pair at each step of the training. This outline stochasticity imposes further advantages on our mechanism. As we are not settling for any static shifting, we are not getting any intertwined directional smoothing during inference. Moreover, our network gets a gradient for multiple online shifting and as a result, our network is tuned to reach a wider minimum compared to the static shifting case. It's been widely accepted by researchers those wide minima help generalization widely.

Apart from that, equation 7 shows that the network tries to equalize two different observations from the same image for two similar noisy observations. As a result, the network is continually back-propagating with respect to the unmodified noisy image which allows propagating underlying clean pixel information along with the surface noise. In processes, the network learns to push back the possible necessary details from the second part of the loss function, while the first aims to suppress the noise to reach for possible smoother.

In essence, our first server as the ‘push' function by introducing necessary smoothing for the task. In contrast, our network part serves as the ‘pull’ function by introducing stochastic identity regularization. Related quantitative and visual demonstrations are present in the following sections and the training framework is shown in Algorithm: Push and Pull.

\begin{figure}[htbp]
\begin{center}
\resizebox{\columnwidth}{!}{%
\begin{tabular}{l l l l l}

    \begin{minipage}{.1\textwidth}
      \includegraphics[width=20mm, height=20mm]{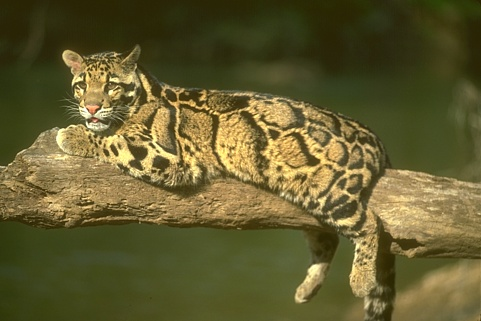}     
         \centering \tiny \textbf{Ground  }\\ \textbf{Truth}
    \label{bsd1}
    \end{minipage}
    
    &
    \begin{minipage}{.1\textwidth}
      \includegraphics[width=20mm, height=20mm]{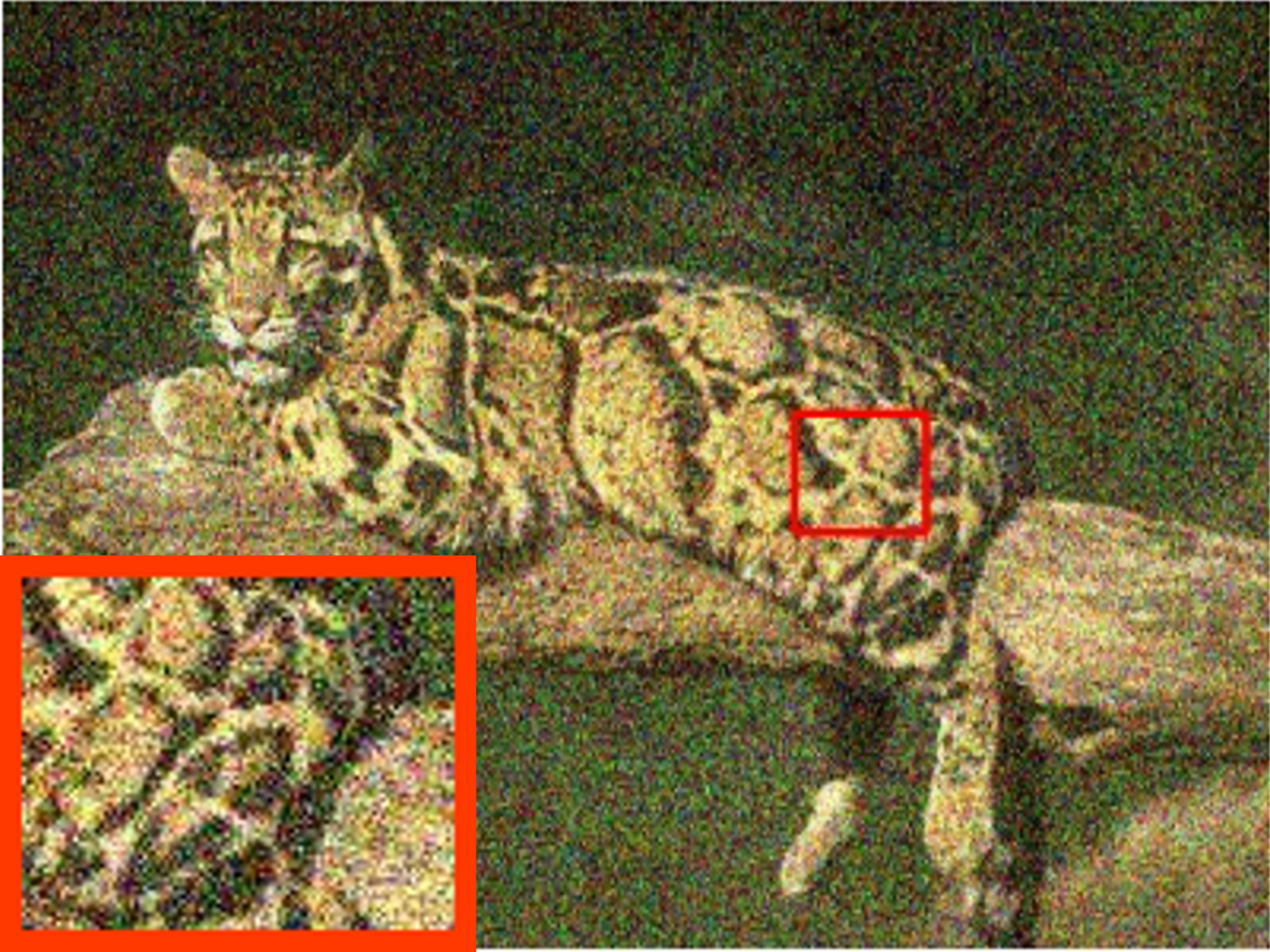}
        \centering \tiny \textbf{Noisy patch}\\ \textbf{15.11 dB}
        \label{bsd2}
    \end{minipage}
    &
    \begin{minipage}{.1\textwidth}
      \includegraphics[width=20mm, height=20mm]{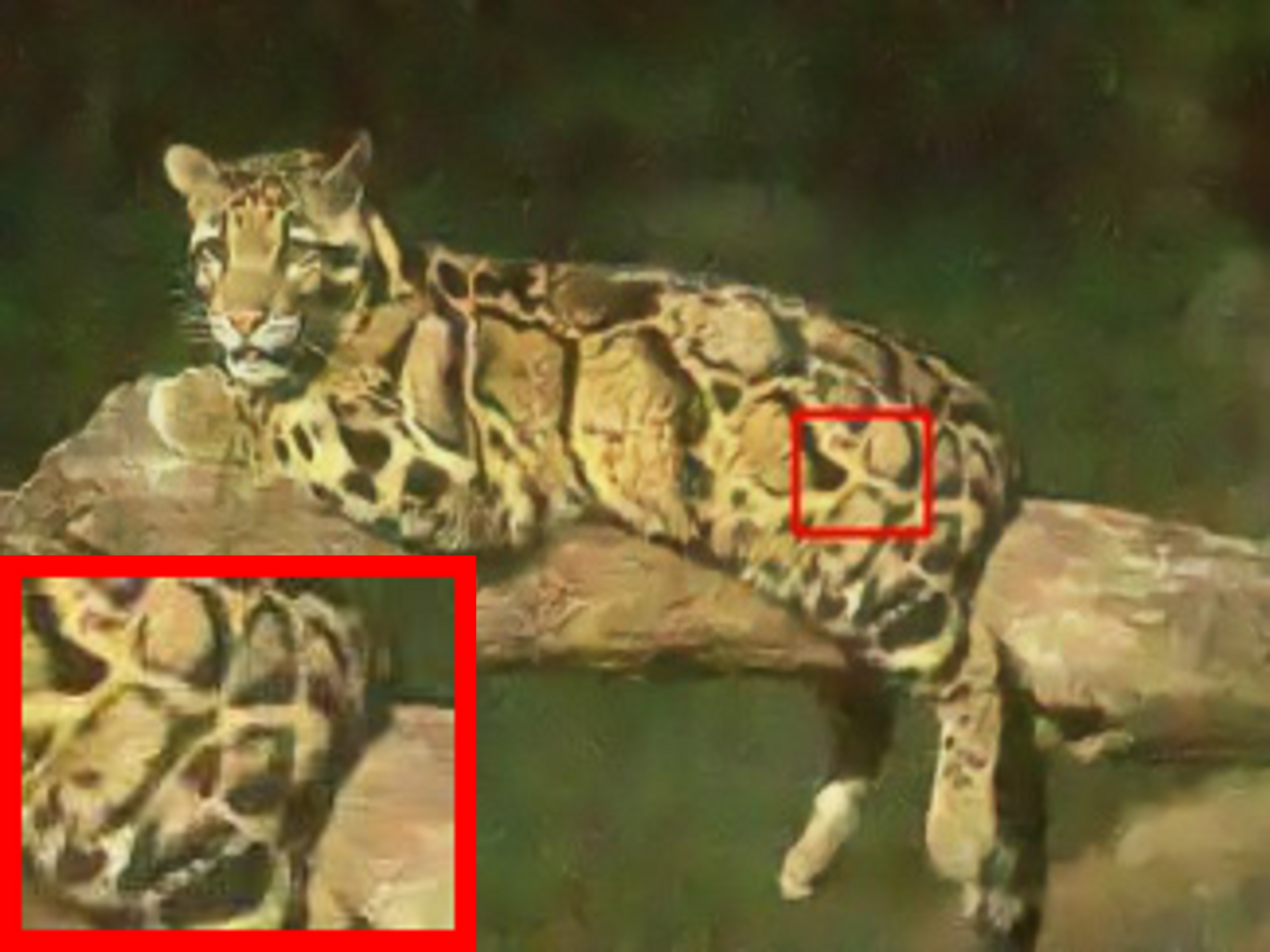}
        \centering \tiny \textbf{DNCNN}\\ \textbf{26.77 dB}
        \label{bsd3}
    \end{minipage}
  &
    \begin{minipage}{.1\textwidth}
     \includegraphics[width=20mm, height=20mm]{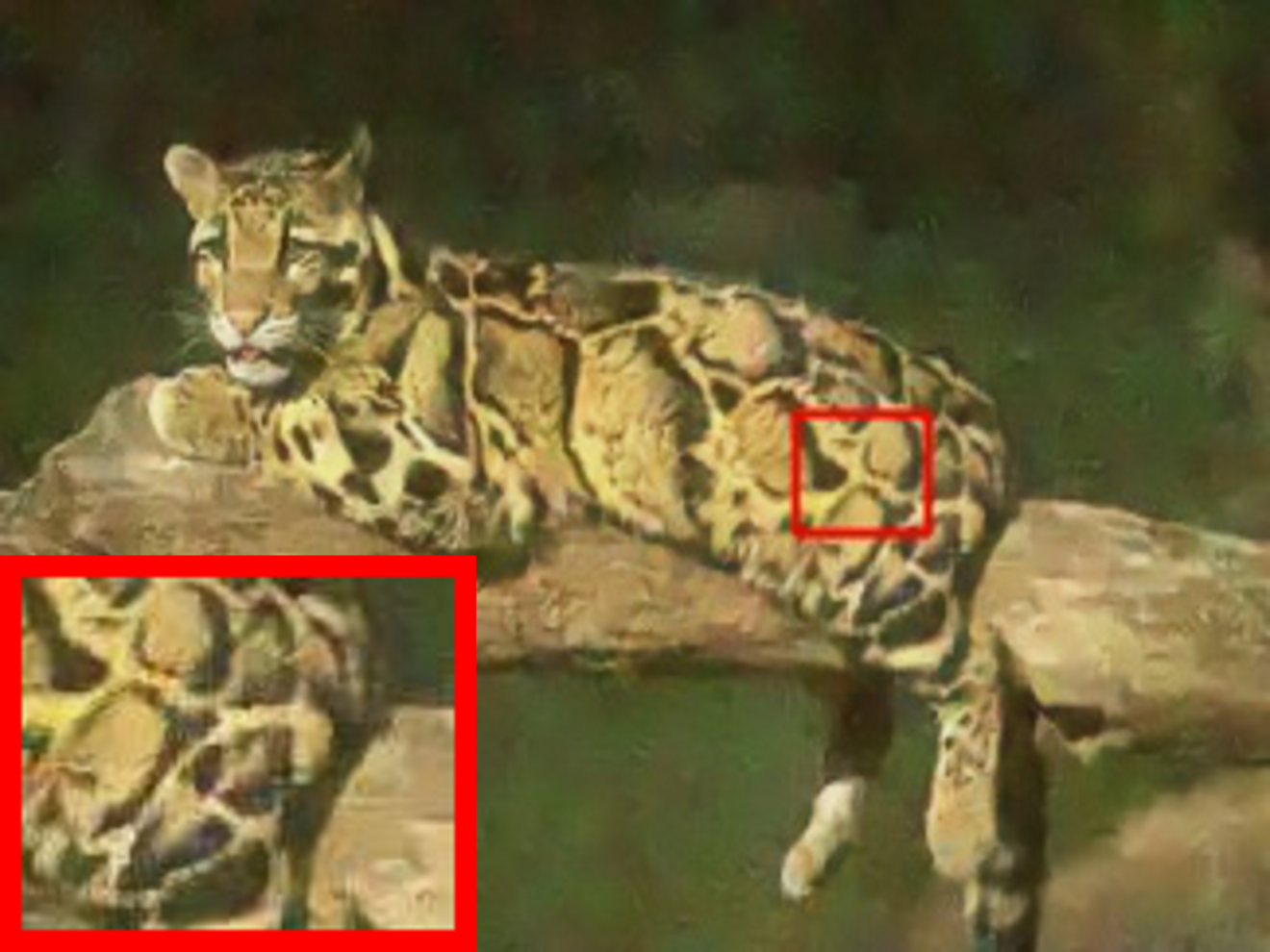}
        \centering \tiny \textbf{N2N}\\ \textbf{26.25 dB}
        \label{bsd4}
    \end{minipage}
    &
    \begin{minipage}{.1\textwidth}
     \includegraphics[width=20mm, height=20mm]{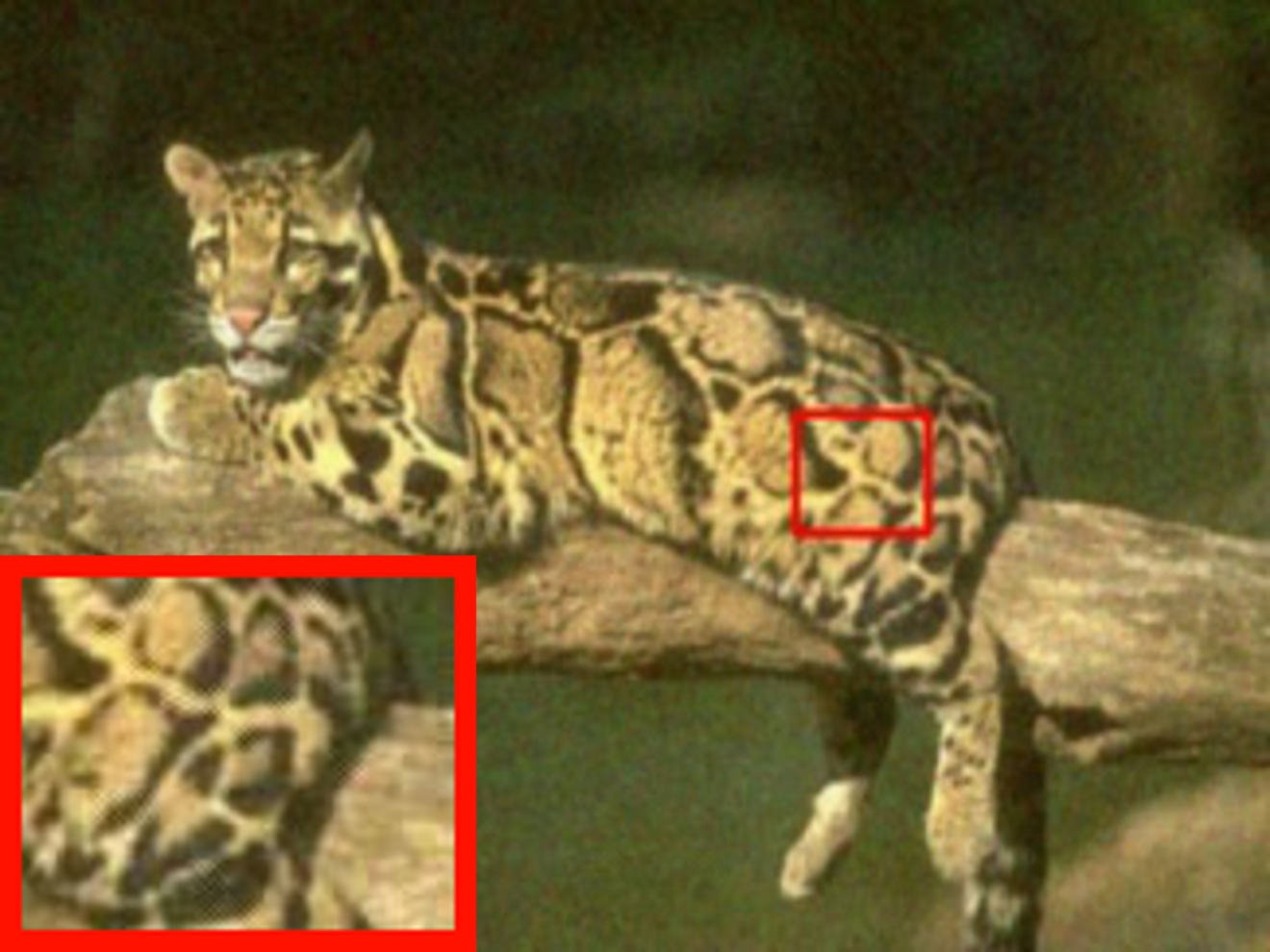}
        \centering \tiny \textbf{N2S}\\ \textbf{26.62 dB}
        \label{bsd5}
    \end{minipage}\\
    \addlinespace
     \begin{minipage}{.1\textwidth}
     \includegraphics[width=20mm, height=20mm]{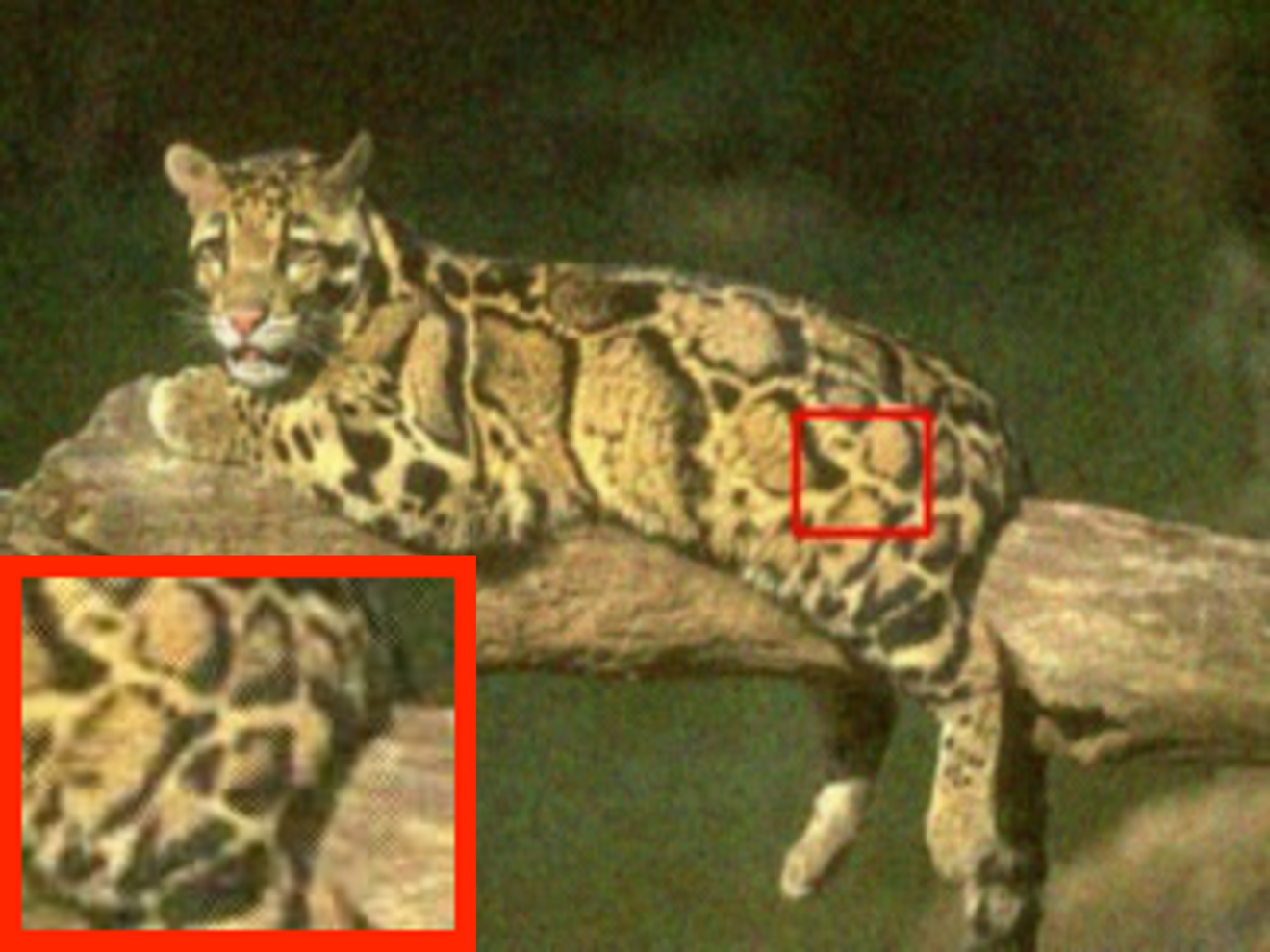}
        \centering \tiny \textbf{N2V}\\ \textbf{25.67 dB}
        \label{bsd6}
    \end{minipage}
    &
     \begin{minipage}{.1\textwidth}
     \includegraphics[width=20mm, height=20mm]{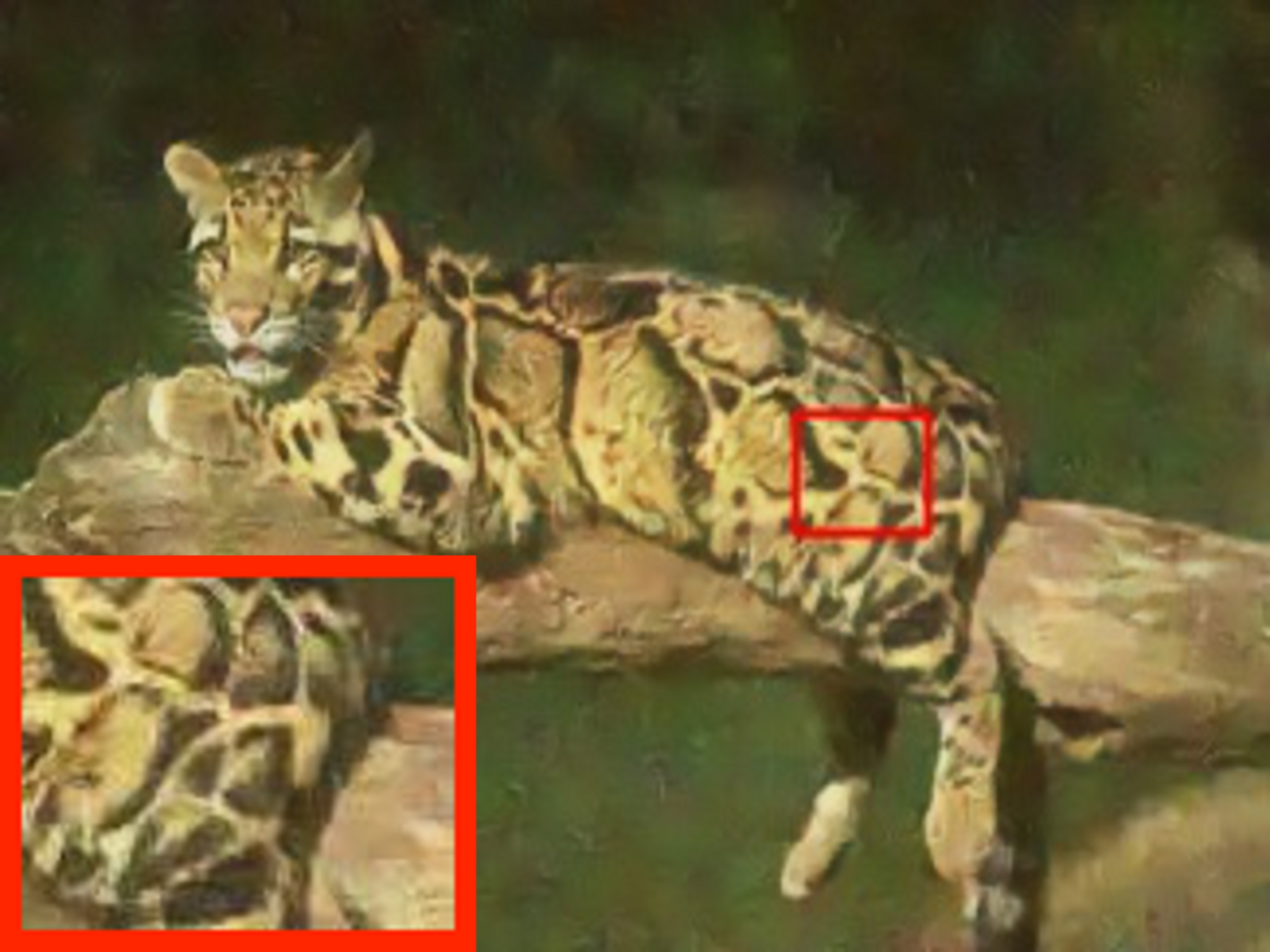}
        \centering \tiny \textbf{Nr2N}\\ \textbf{26.13 dB}
        \label{bsd7}
    \end{minipage}
    &
     \begin{minipage}{.1\textwidth}
     \includegraphics[width=20mm, height=20mm]{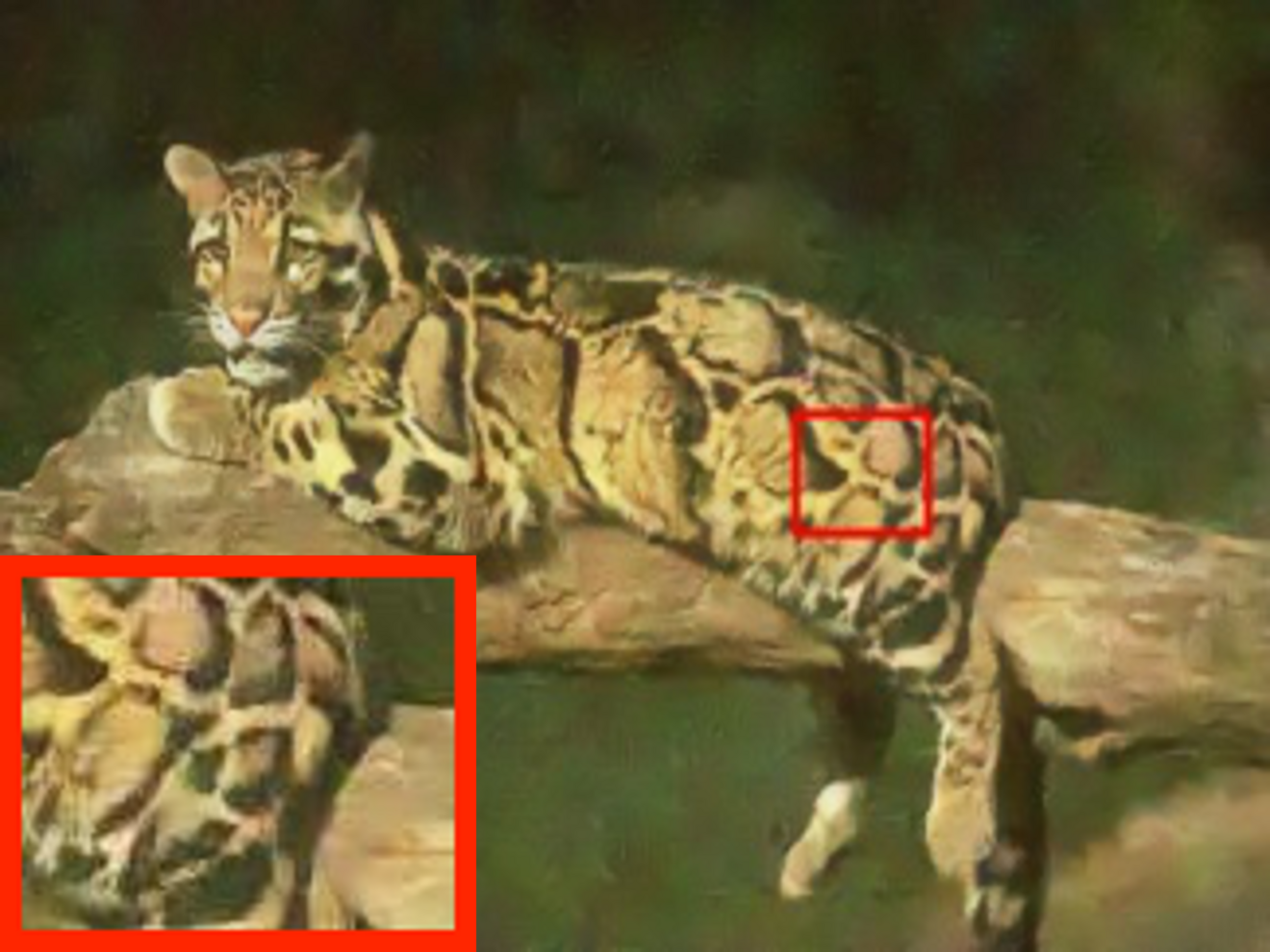}
        \centering \tiny \textbf{R2R}\\ \textbf{26.69 dB}
        \label{bsd8}
    \end{minipage}
    &
     \begin{minipage}{.1\textwidth}
     \includegraphics[width=20mm, height=20mm]{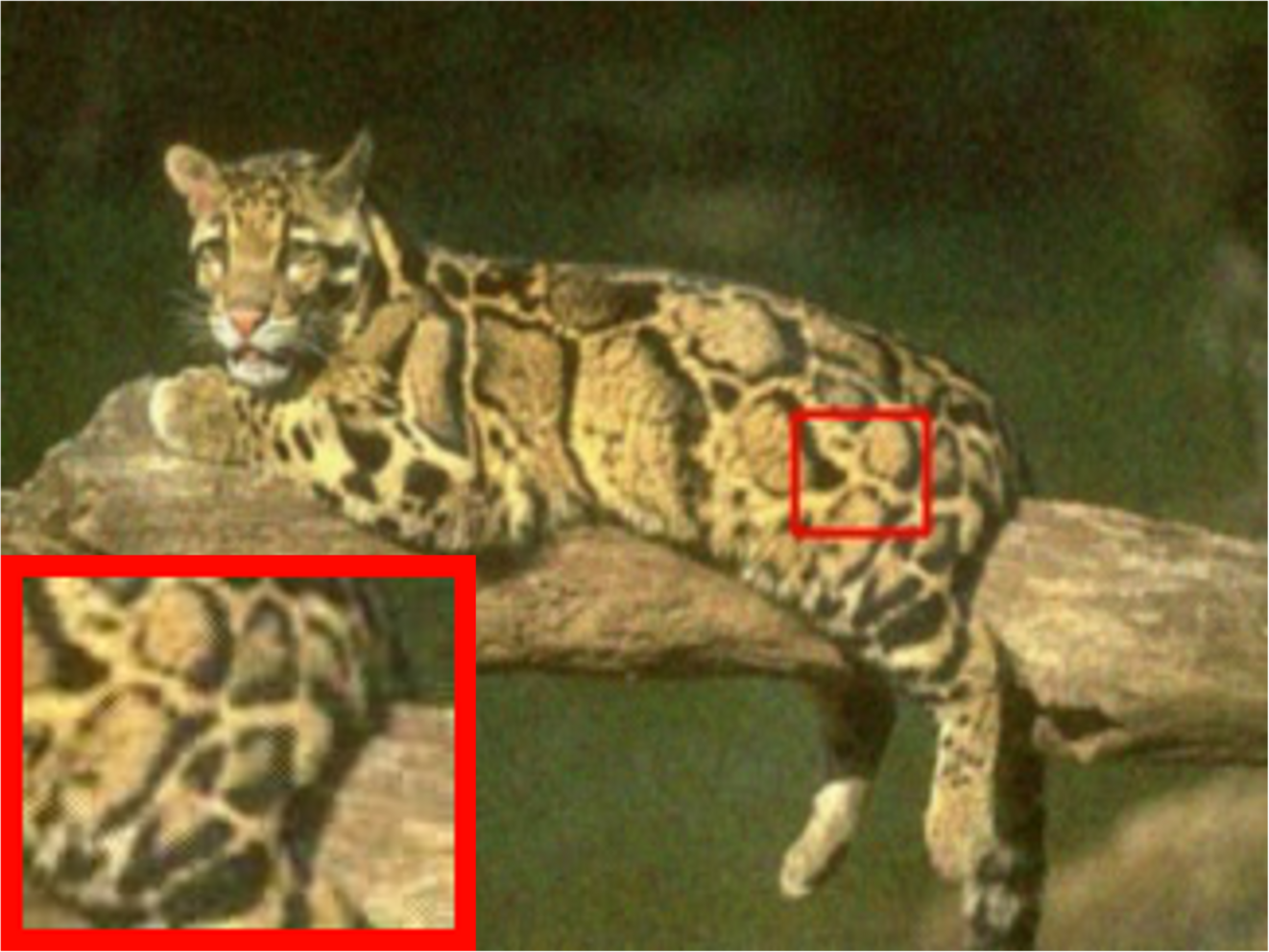}
        \centering \tiny \textbf{SURE}\\ \textbf{26.54 dB}
        \label{bsd9}
    \end{minipage}
    &
     \begin{minipage}{.1\textwidth}
     \includegraphics[width=20mm, height=20mm]{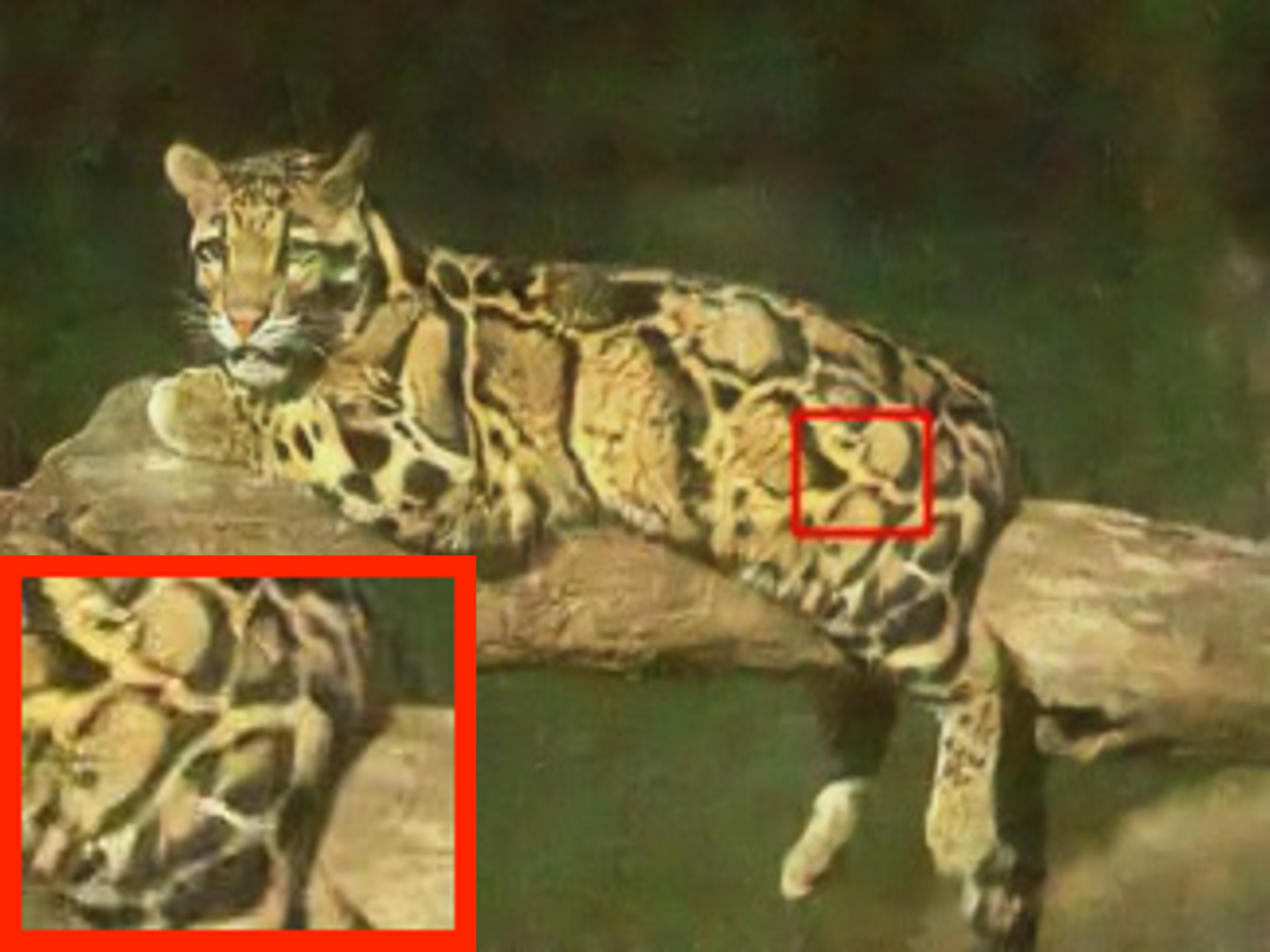}
        \centering \tiny \textbf{Proposed}\\ \textbf{26.81 dB}
        \label{bsd10}
    \end{minipage}\\

\end{tabular}
}
 
\caption{Visual quality comparison on the BSD 68 dataset with AWGN noise level $\sigma = 50$ (zoom-in for best view).}
\label{ BSD AWGN}
 
\end{center}

\end{figure}
\begin{figure}[htbp]
\begin{center}
\resizebox{\columnwidth}{!}{%
\begin{tabular}{l l l l l}

    \begin{minipage}{.1\textwidth}
      \includegraphics[width=20mm, height=20mm]{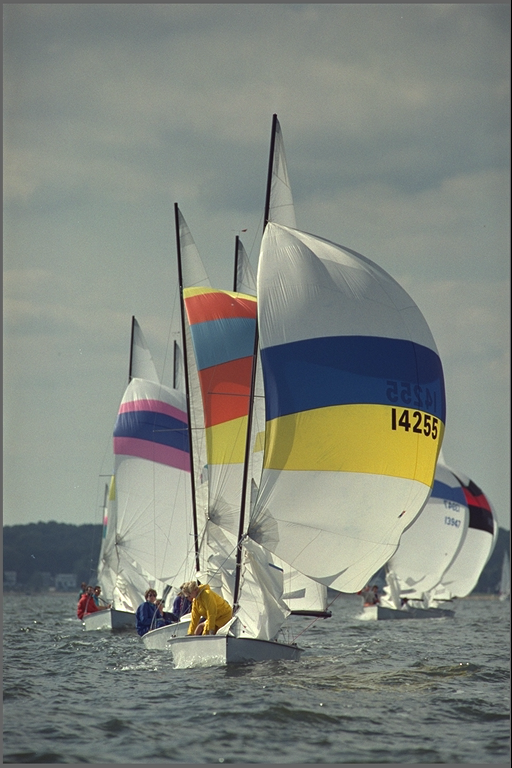}     
         \centering \tiny \textbf{Ground  }\\ \textbf{Truth}
    \label{KODAK1}
    \end{minipage}
    
    &
    \begin{minipage}{.1\textwidth}
      \includegraphics[width=20mm, height=20mm]{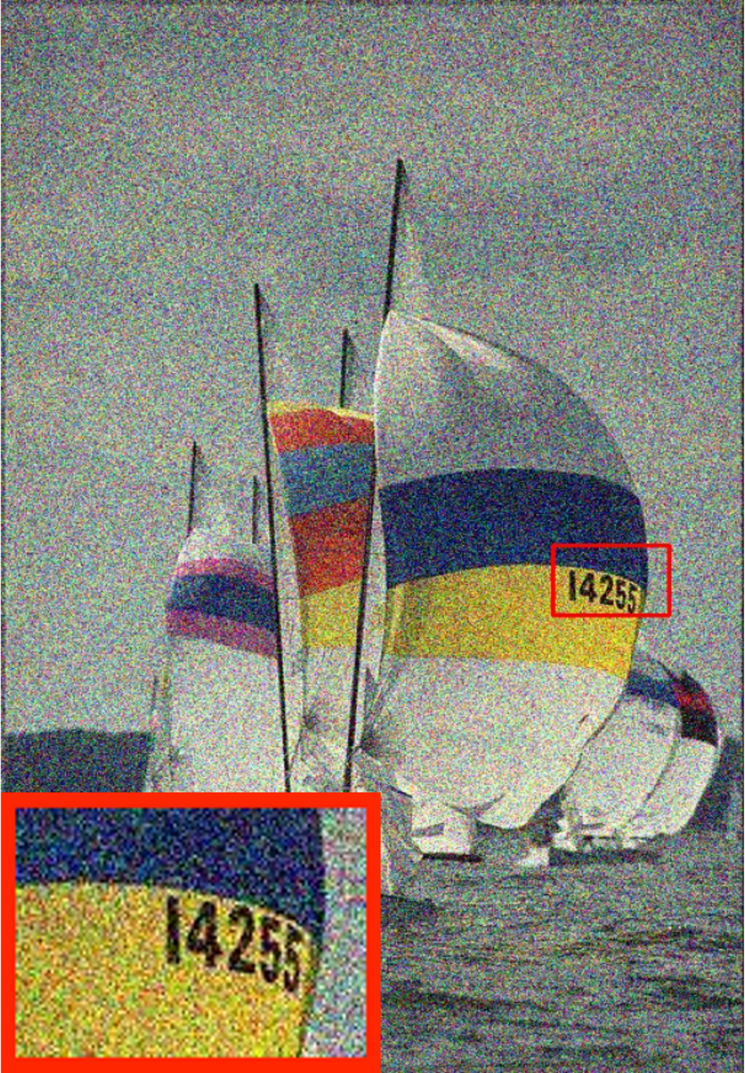}
        \centering \tiny \textbf{Noisy patch}\\ \textbf{14.46 dB}
        \label{KODAK2}
    \end{minipage}
    &
    \begin{minipage}{.1\textwidth}
      \includegraphics[width=20mm, height=20mm]{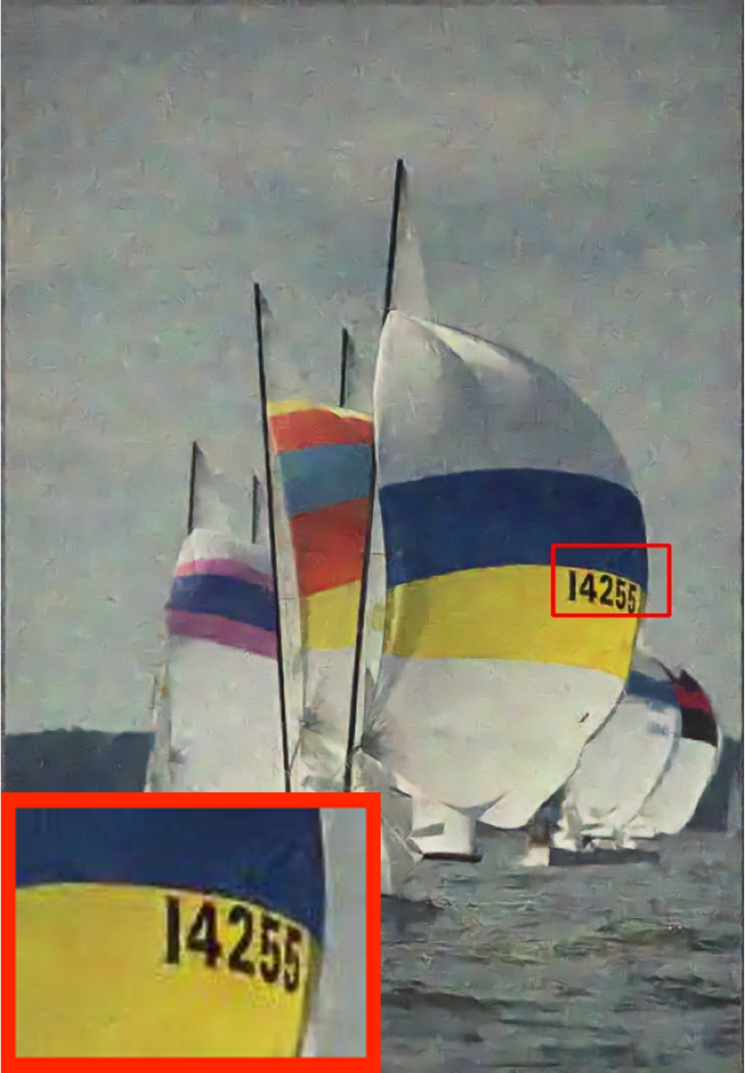}
        \centering \tiny \textbf{DNCNN}\\ \textbf{29.35 dB}
        \label{KODAK3}
    \end{minipage}
  &
    \begin{minipage}{.1\textwidth}
     \includegraphics[width=20mm, height=20mm]{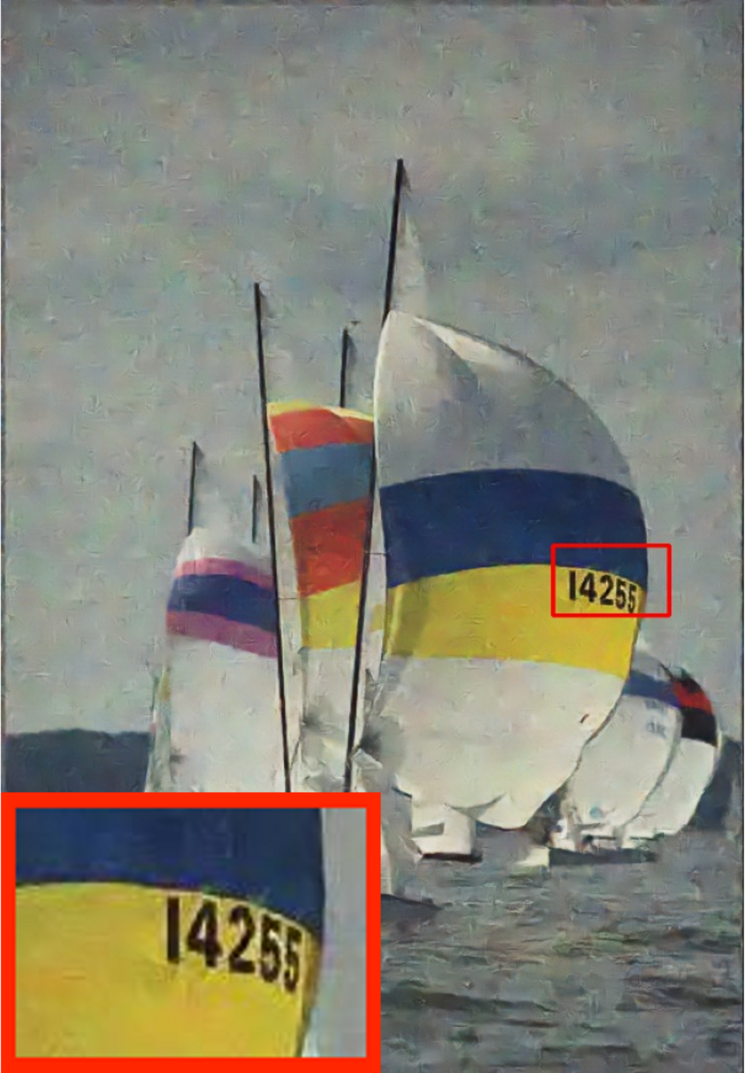}
        \centering  \tiny \textbf{N2N}\\ \textbf{28.88 dB}
        \label{KODAK4}
    \end{minipage}
    &
    \begin{minipage}{.1\textwidth}
     \includegraphics[width=20mm, height=20mm]{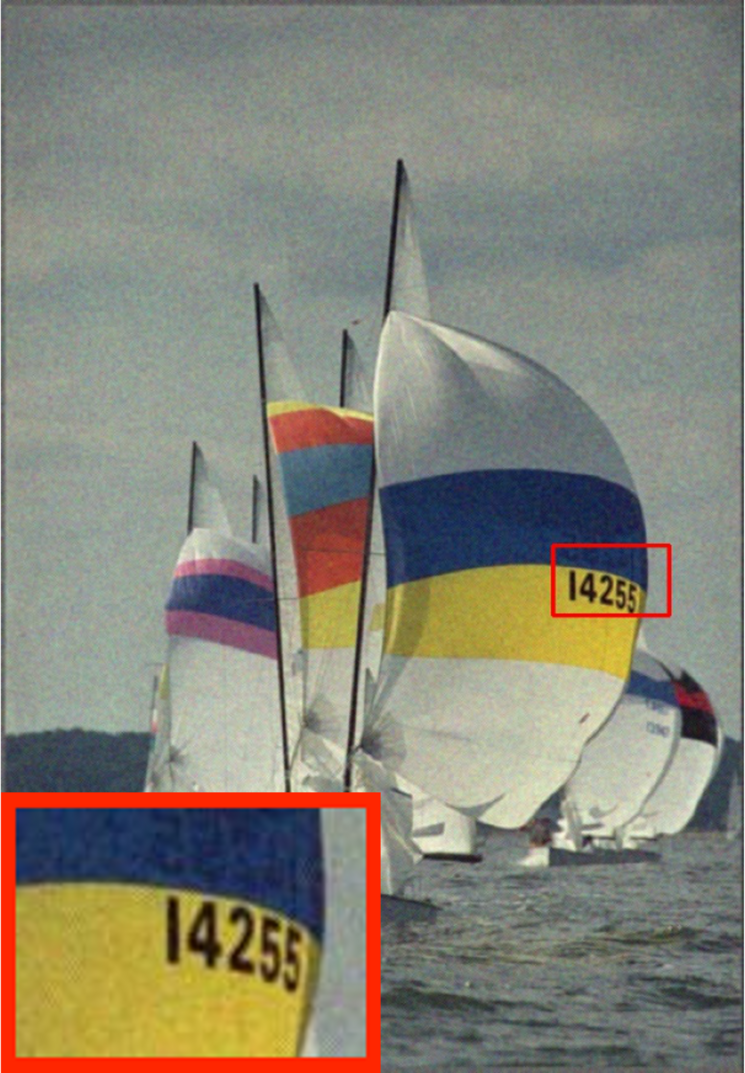}
        \centering  \tiny \textbf{N2S}\\ \textbf{28.03 dB}
        \label{KODAK5}
    \end{minipage}\\
    \addlinespace
     \begin{minipage}{.1\textwidth}
     \includegraphics[width=20mm, height=20mm]{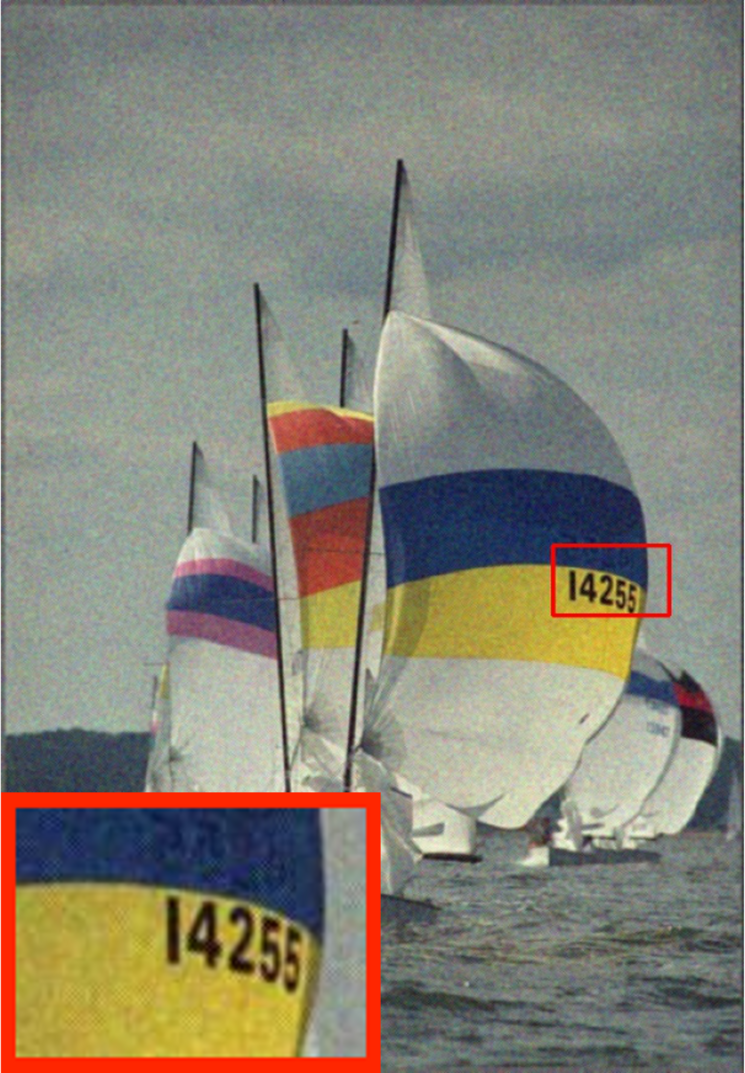}
        \centering  \tiny \textbf{N2V}\\ \textbf{26.69 dB}
        \label{KODAK6}
    \end{minipage}
    &
     \begin{minipage}{.1\textwidth}
     \includegraphics[width=20mm, height=20mm]{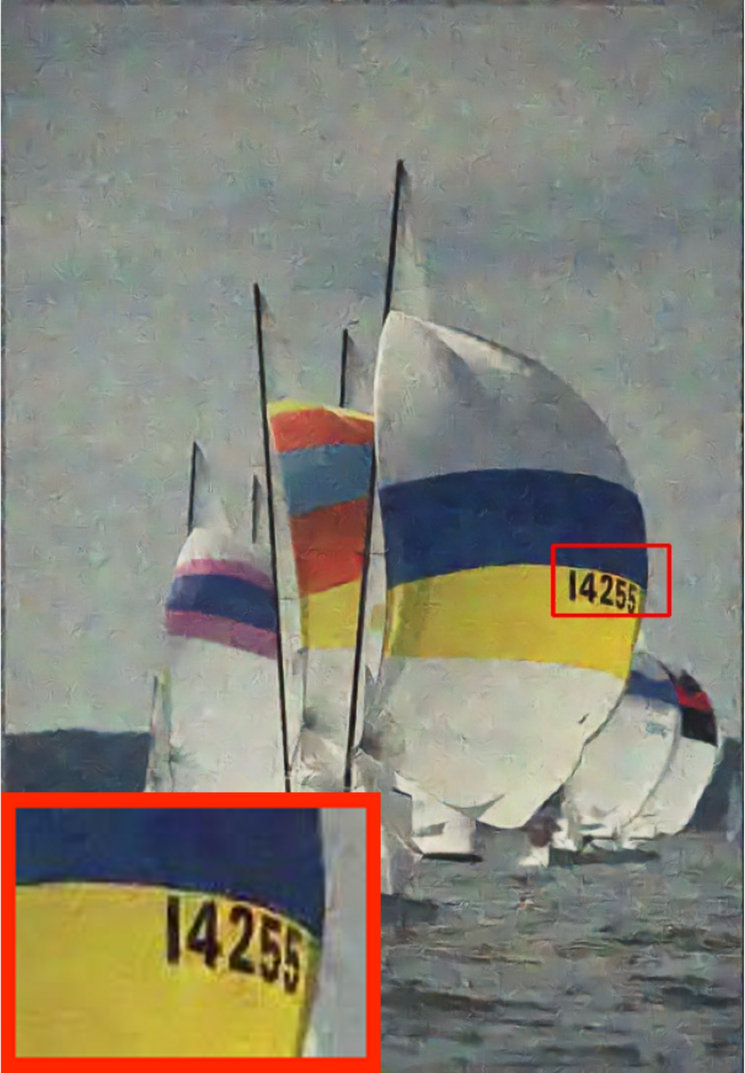}
        \centering  \tiny \textbf{Nr2N}\\ \textbf{28.71 dB}
        \label{KODAK7}
    \end{minipage}
    &
     \begin{minipage}{.1\textwidth}
     \includegraphics[width=20mm, height=20mm]{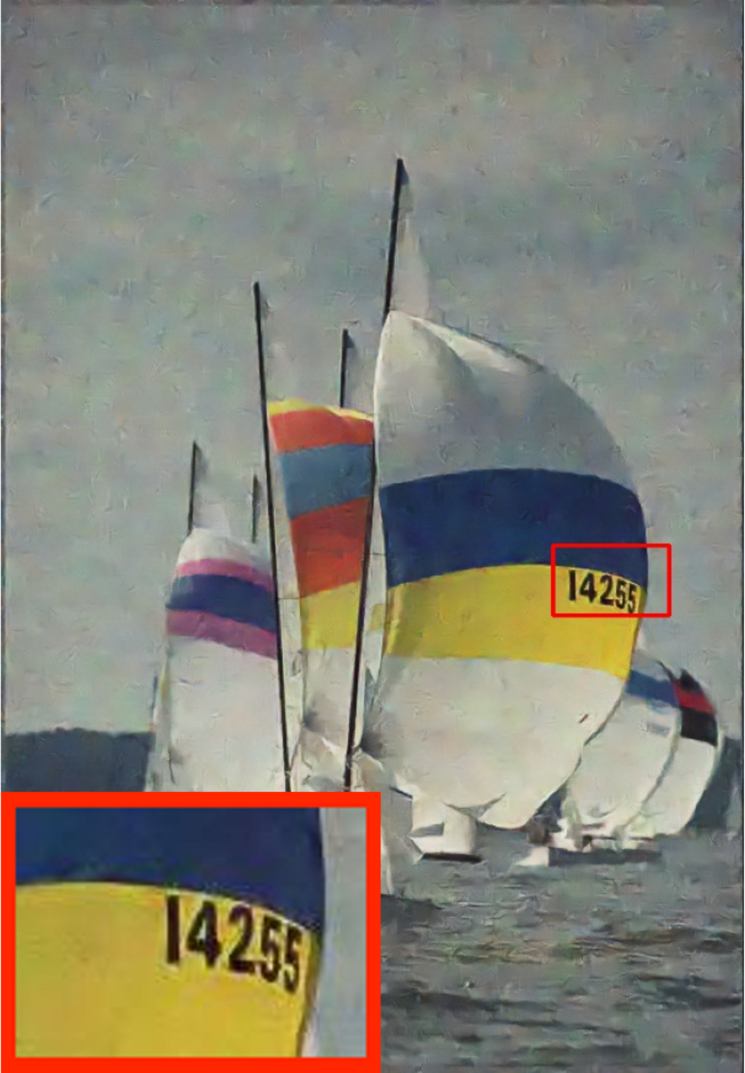}
        \centering  \tiny \textbf{R2R}\\ \textbf{29.34 dB}
        \label{KODAK8}
    \end{minipage}
    &
     \begin{minipage}{.1\textwidth}
     \includegraphics[width=20mm, height=20mm]{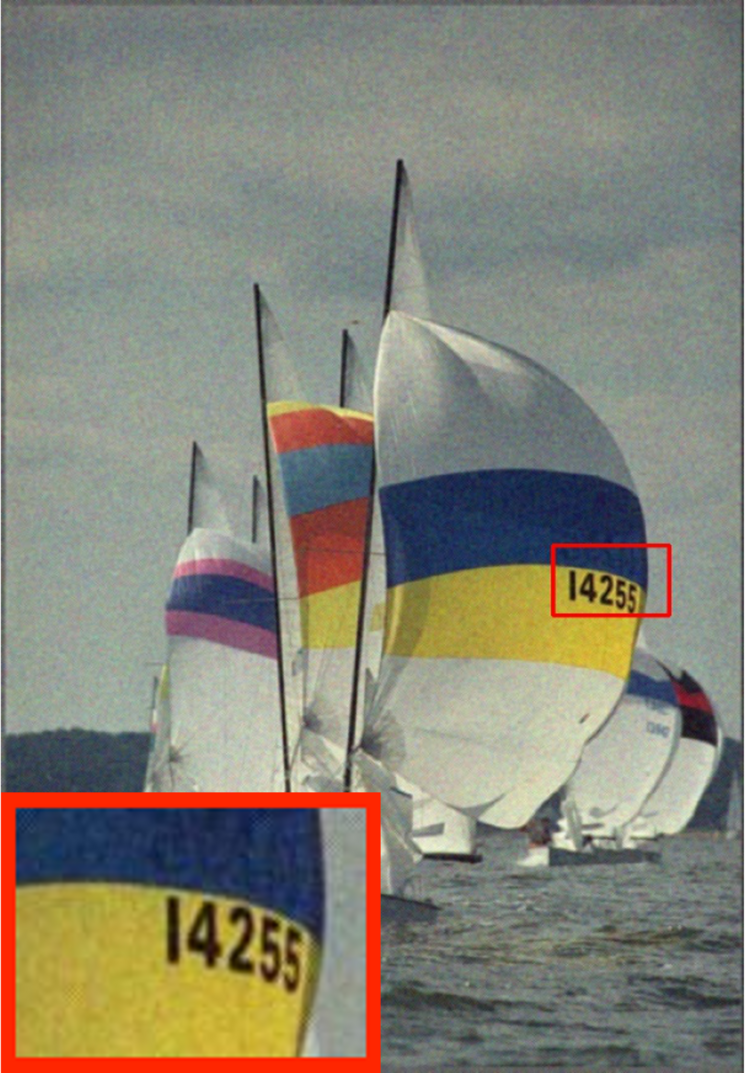}
        \centering  \tiny \textbf{SURE}\\\textbf{27.89 dB} 
        \label{KODAK9}
    \end{minipage}
    &
     \begin{minipage}{.1\textwidth}
     \includegraphics[width=20mm, height=20mm]{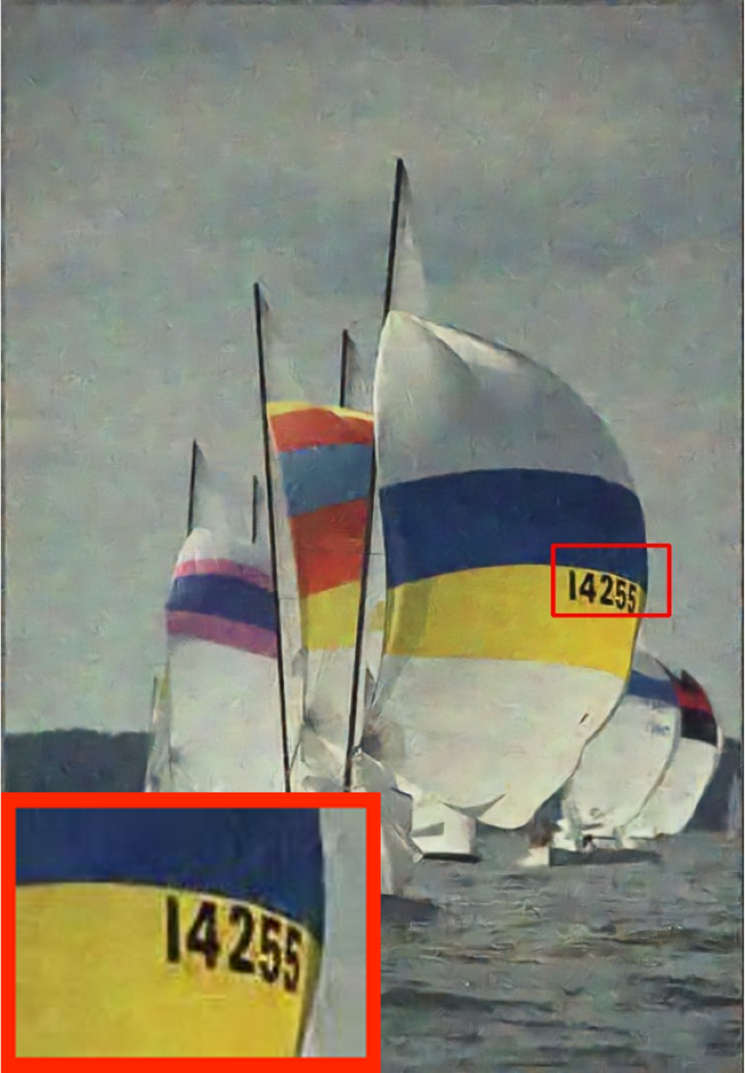}
        \centering  \tiny \textbf{Proposed}\\\textbf{29.66 dB}
        \label{KODAK10}
    \end{minipage}\\

\end{tabular}
}
 
\caption{Visual quality comparison on the KODAK dataset with AWGN noise level $\sigma = 50$ (zoom-in for best view).}
\label{  KODAK AWGN}
  
\end{center}

\end{figure}


\section{Experimental Results}
\section{Experiments}
The main objective of the experiments is to manifest that, under different conditions, the proposed noise removal technique provides positive results in denoising performance. In this section, we evaluate our proposed  training algorithm on Additive Gaussian Noise (AWGN) and Poisson noise removal. In our experiments, we employ the same architecture as that of DnCNN \cite{zhang2017beyond} which is a baseline for denoising DNN. The results of the compared methods are either cited directly from the literature or by using the pre-trained models or the codes provided by the authors. For the recorrupted2recorrupted \cite{pang2021recorrupted} study, we have to infer the the test images according to the red, green, and blue channel, followed by mandatory merging to reconstruct desired view. This is due to the author's specific gray-scale implementation code, as available in the GitHub repository. For the noise2self \cite{batson2019noise2self} study, we utilized the available scipy implementation to obtain the result.

\begin{center}
\begin{table*}[ht]
\setlength{\tabcolsep}{1.8pt}
\newcommand\setrow[1]{\gdef\rowmac{#1}#1\ignorespaces}
\caption{Average  SSIM/PSNR (DB) results on  different dataset corrupted by additive gaussian noise with $\sigma =25$ and $\sigma =50$  }
\centering
\begin{tabular}{@{\extracolsep{1pt}}llccccccccccc}
\toprule   
{}  & \multicolumn{3}{c}{$\sigma=25$}  & \multicolumn{3}{c}{$\sigma=50$}\\
 \cmidrule{2-4} 
 \cmidrule{5-7} 
Method  & BSD 68 & Kodak & UR100 & BSD 68 & Kodak & UR100  \\ 
\midrule
DnCNN \cite{zhang2017beyond} & 0.9024/29.21   & 0.9073/30.02 & 0.9128/28.35 & 0.8338/26.23 & 0.8497/27.14 & 0.8435/25.01\\
N2S \cite{batson2019noise2self}  & 0.8856/28.21	& 0.9035/29.24 & 0.8723/25.56 & 0.7986/26.01 & 0.8015/26.65 & 0.7985/24.05\\
N2V\cite{krull2019noise2void}	   & 0.8751/27.88   & 0.8898/28.79 & 0.8601/25.26 & 0.7532/25.09 & 0.7499/25.63 & 0.7630/23.41\\
N2N \cite{lehtinen2018noise2noise}	   & 0.8935/28.84	& 0.8997/29.67 & 0.9058/27.99 & 0.8213/25.74 & 0.8390/26.71 & 0.8336/24.60\\
Nr2N\cite{moran2020noisier2noise}   & 0.8856/28.52	& 0.8935/29.39 & 0.8993/27.68 & 0.8189/25.64 & 0.8362/26.61 & 0.8310/24.50\\
R2R \cite{pang2021recorrupted}   & 0.9012/29.16	& 0.9064/29.98 & 0.9119/28.31 & 0.8315/26.14 & 0.8490/27.12 & 0.8432/25.00\\
SURE \cite{soltanayev2018training}  & 0.8848/28.18	& 0.9023/29.20 & 0.8711/25.54 & 0.7942/25.92 & 0.7966/26.55 & 0.7950/23.98\\
\bottomrule
\textbf{Proposed}  &  \textbf{0.9152/29.97} & \textbf{0.9183/30.81} &\textbf{ 0.9245/29.29} & \textbf{0.8416/26.29} & \textbf{0.8591/27.34} & \textbf{0.8546/25.28}\\
\bottomrule
\end{tabular}
\label{Tab:t1}
\end{table*}
\end{center}

\begin{center}
\begin{table*}[ht]
\setlength{\tabcolsep}{1.8pt}
\newcommand\setrow[1]{\gdef\rowmac{#1}#1\ignorespaces}
\caption{Average  SSIM/PSNR (DB) results on  different datasets corrupted by  Poisson noise with levels $\lambda \in[5, 50]$}
\centering
\begin{tabular}{@{\extracolsep{1pt}}llccccccccccc}
\toprule   
{}  & \multicolumn{3}{c}{$\lambda \in[5, 50]$}  \\
 \cmidrule{2-4}  
Method  & BSD 68 & Kodak & UR100  \\ 
\midrule
DnCNN \cite{zhang2017beyond} & 0.9352/30.81   & 0.9242/30.74 & 0.9360/29.23 \\
N2S \cite{batson2019noise2self}  & 0.8915/27.97	& 0.8978/28.81 & 0.9019/26.59 \\
N2V\cite{krull2019noise2void}	 & 0.8731/27.13  & 0.8834/27.96 & 0.9049/26.70 \\
N2N \cite{lehtinen2018noise2noise}& 0.9234/30.19 & 0.9119/30.17 & 0.9329/28.95 \\
Nr2N\cite{moran2020noisier2noise} & 0.8876/27.88 & 0.8958/29.13 & 0.9090/27.09 \\
R2R \cite{pang2021recorrupted}   & 0.9212/29.89	&  0.9115/29.91 & 0.9297/28.72\\ 
SURE \cite{soltanayev2018training} & 0.9011/28.43 & 0.9105/29.67 & 0.9154/27.33 \\
\bottomrule
\textbf{Proposed}  &  \textbf{0.9256/30.00} & \textbf{0.9232/30.18} &\textbf{ 0.9332/28.95}\\
\bottomrule
\end{tabular}
\label{Tab:t2}
\end{table*}
\end{center}



\subsection{Training Details}
For this work, we followed a standard training strategy. To start with, we created the necessary observations required for the training as per our algorithm. To its extent, we provided a diverse synthetic sample to our data stream to train a network for interesting images with different sets of noisy observations. For training, we maintained batch size as 16 and standard data augmentations. In addition to that,  we run our setup for 400 epochs and scheduled the learning rate according to the epochs.  We converted our input patches with the size of 40. We have only used the Div2k dataset for final training. Additionally, we tried training with MIT 5k, BSD 500, and GLADNet datasets and achieved comparable results.


\begin{figure}[htbp]
\begin{center}
\resizebox{\columnwidth}{!}{%
\begin{tabular}{l l l l l}

    \begin{minipage}{.1\textwidth}
      \includegraphics[width=16mm, height=20mm]{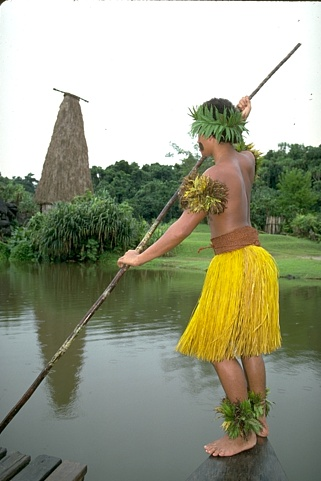}   
         \centering \tiny \textbf{Ground  }\\ \textbf{Truth}
    \label{BSDP1}
    \end{minipage}
    
    &

    \begin{minipage}{.1\textwidth}
      \includegraphics[width=16mm, height=20mm]{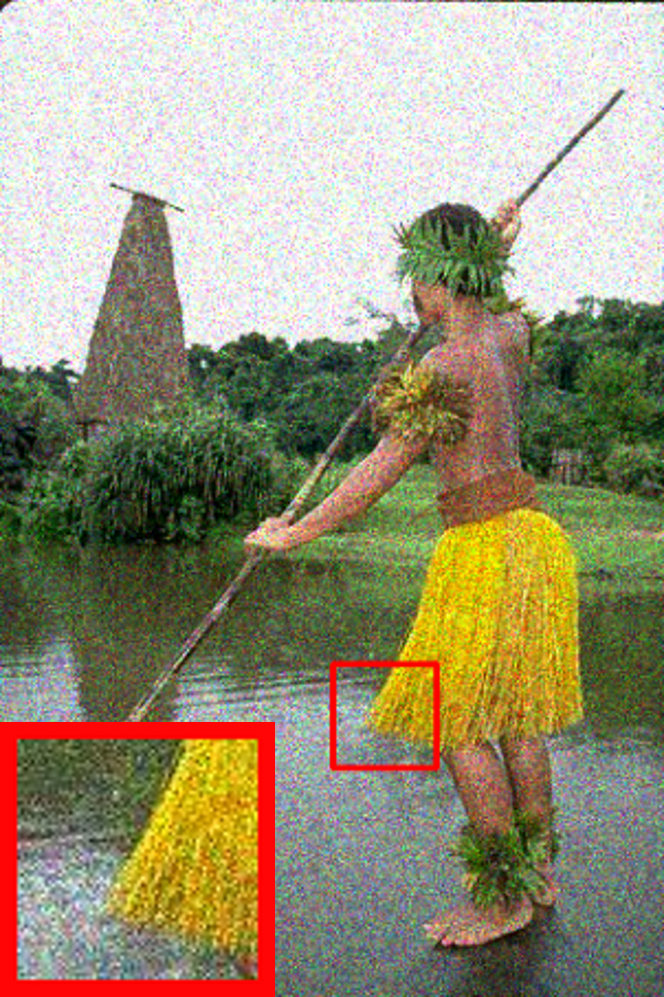}  
         \centering \tiny \textbf{Noisy patch}\\ \textbf{18.67 dB}
    \label{BSDP2}
    \end{minipage}
    &
    \begin{minipage}{.1\textwidth}
      \includegraphics[width=16mm, height=20mm]{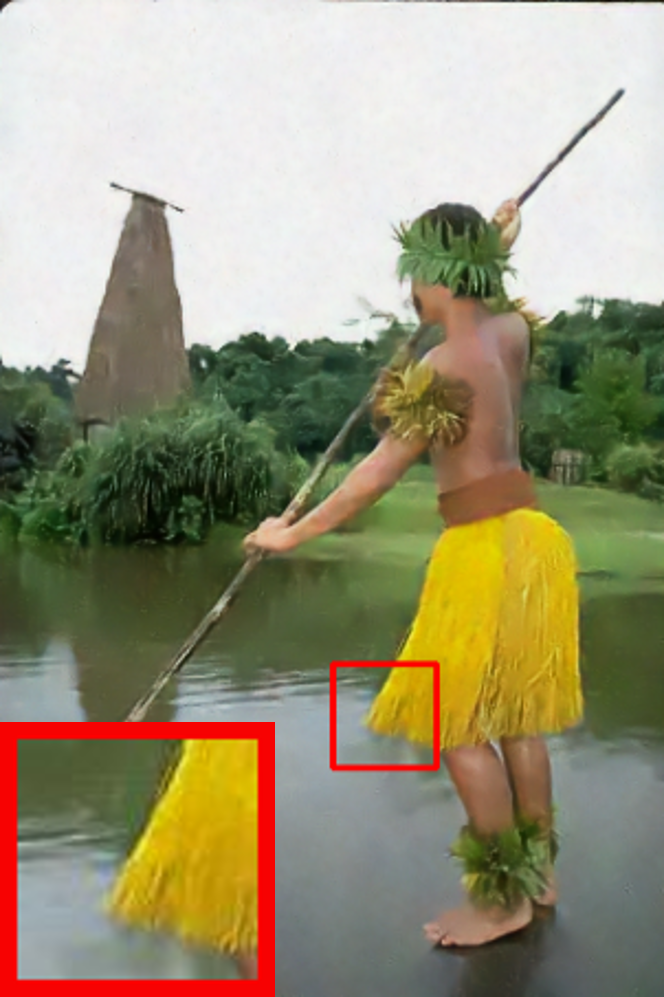}
        \centering \tiny \textbf{DNCNN}\\ \textbf{26.71 dB}
        \label{BSDP3}
    \end{minipage}
  &
    \begin{minipage}{.1\textwidth}
     \includegraphics[width=16mm, height=20mm]{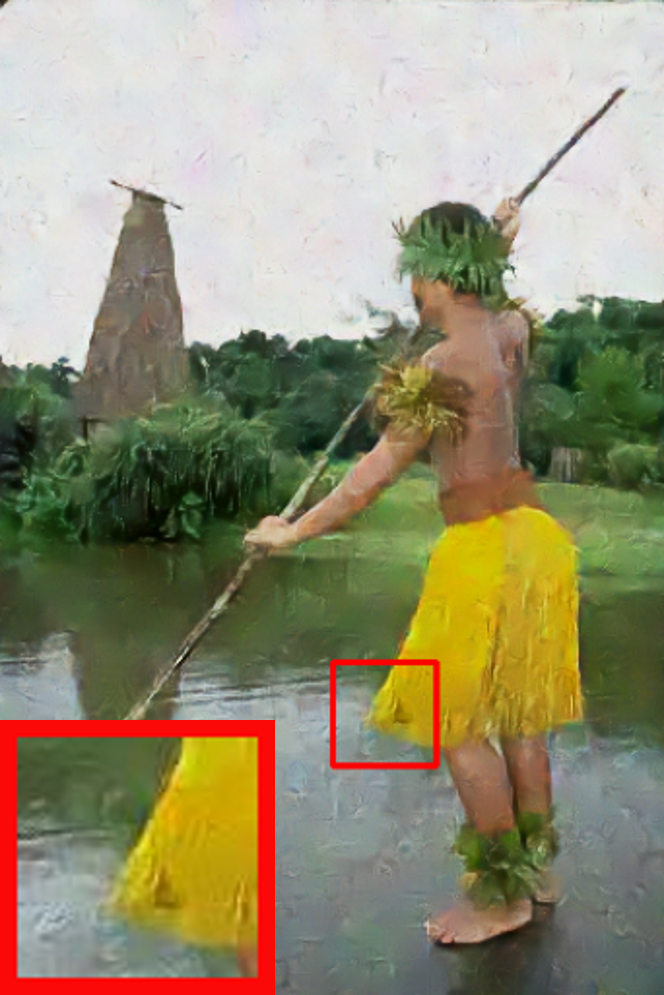}
        \centering  \tiny \textbf{N2N}\\ \textbf{22.50 dB}
        \label{BSDP4}
    \end{minipage}
    &
    \begin{minipage}{.1\textwidth}
     \includegraphics[width=16mm, height=20mm]{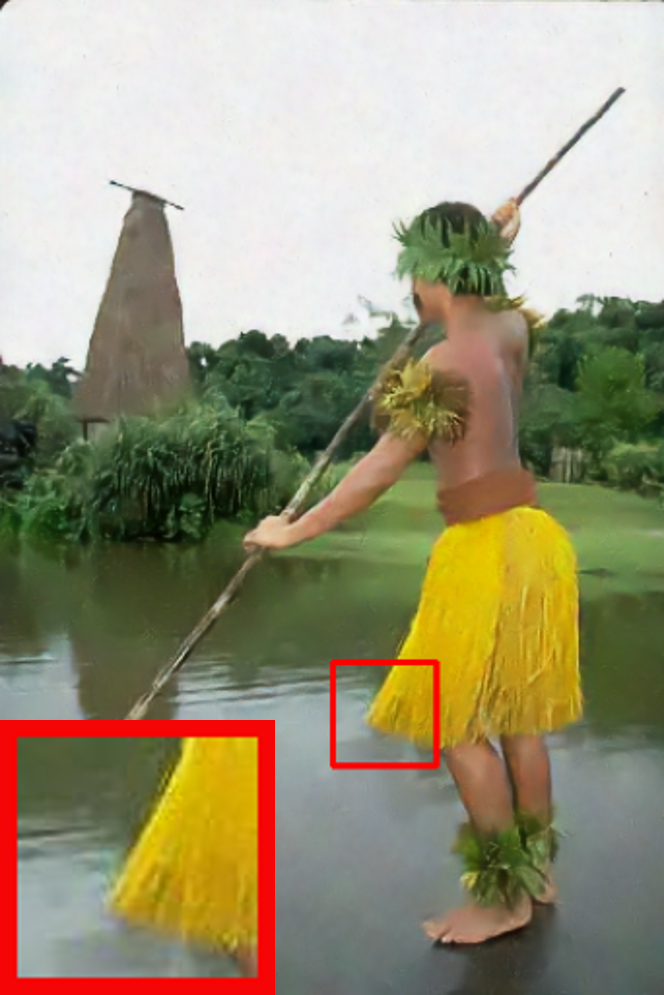}
        \centering  \tiny \textbf{N2S}\\ \textbf{26.62 dB}
        \label{BSDP5}
    \end{minipage}\\
    \addlinespace
     \begin{minipage}{.1\textwidth}
     \includegraphics[width=16mm, height=20mm]{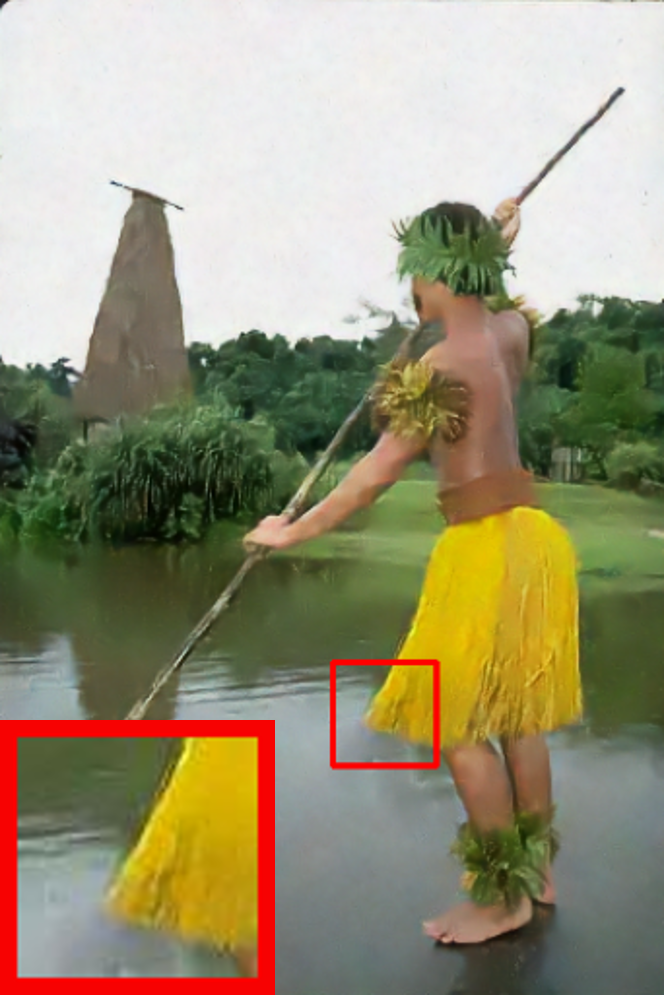}
        \centering  \tiny \textbf{N2V}\\ \textbf{26.31 dB}
        \label{BSDP6}
    \end{minipage}
    &
     \begin{minipage}{.1\textwidth}
     \includegraphics[width=16mm, height=20mm]{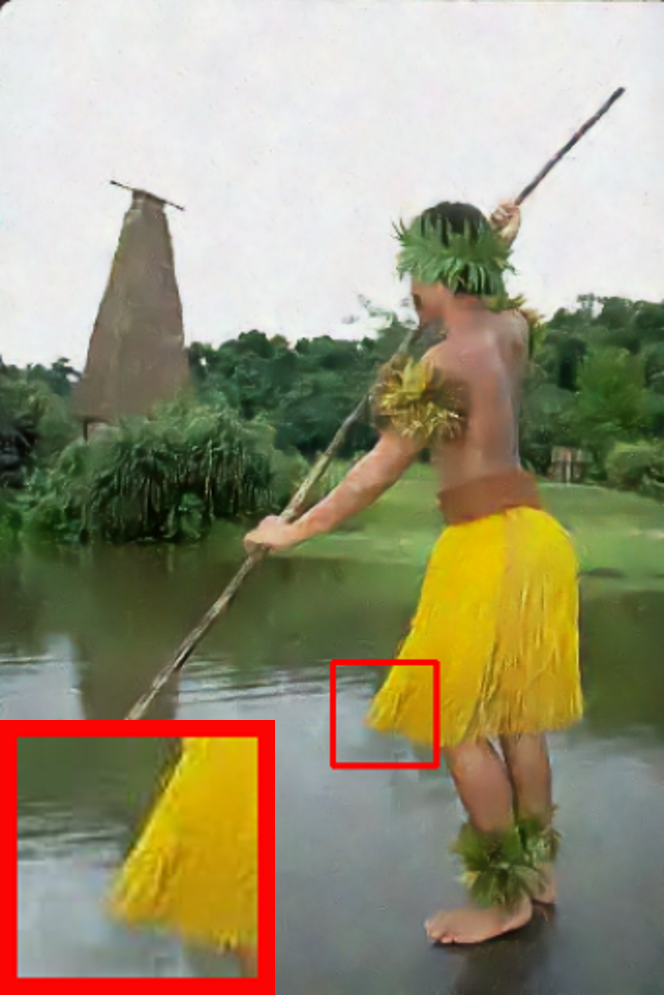}
        \centering  \tiny \textbf{Nr2N}\\ \textbf{25.60 dB}
        \label{BSDP7}
    \end{minipage}
    &
     \begin{minipage}{.1\textwidth}
     \includegraphics[width=16mm, height=20mm]{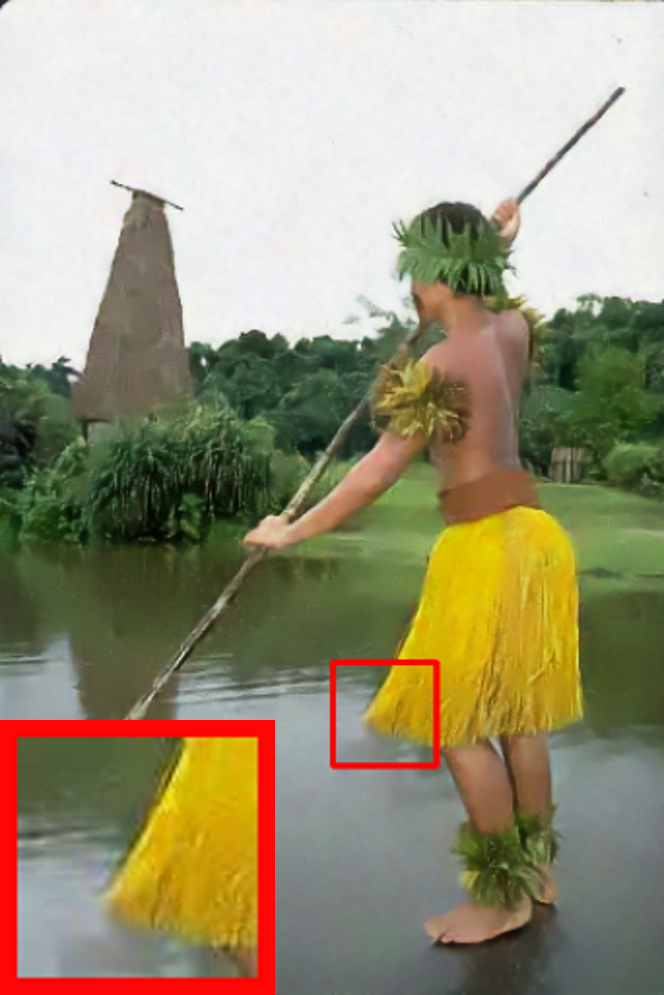}
        \centering  \tiny \textbf{R2R}\\ \textbf{27.07 dB}
        \label{BSDP8}
    \end{minipage}
    &
     \begin{minipage}{.1\textwidth}
     \includegraphics[width=16mm, height=20mm]{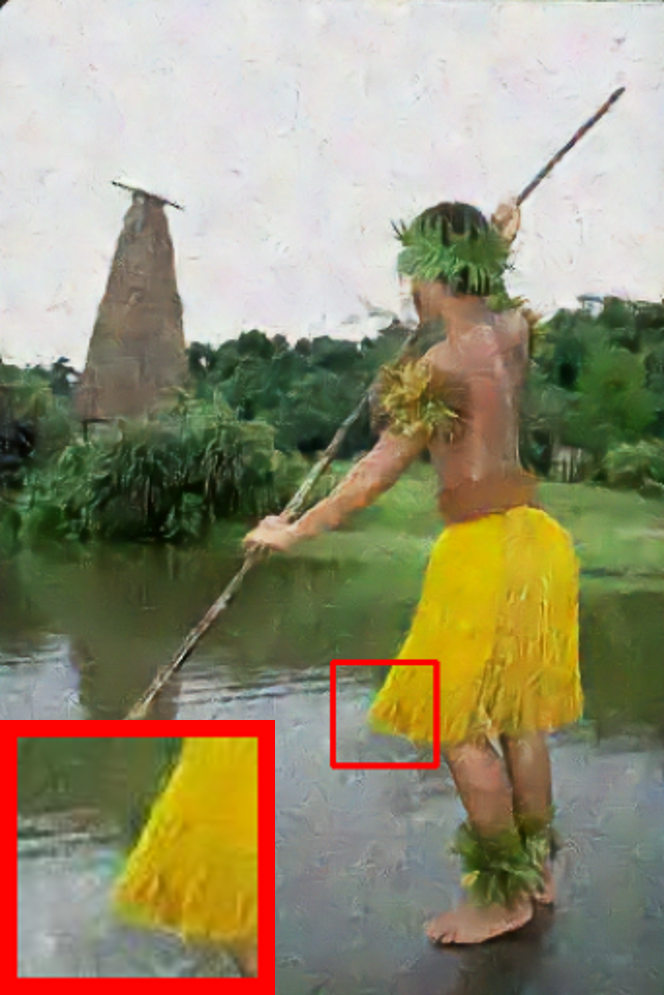}
        \centering  \tiny \textbf{SURE}\\\textbf{22.55 dB} 
        \label{BSDP9}
    \end{minipage}
    &
     \begin{minipage}{.1\textwidth}
     \includegraphics[width=16mm, height=20mm]{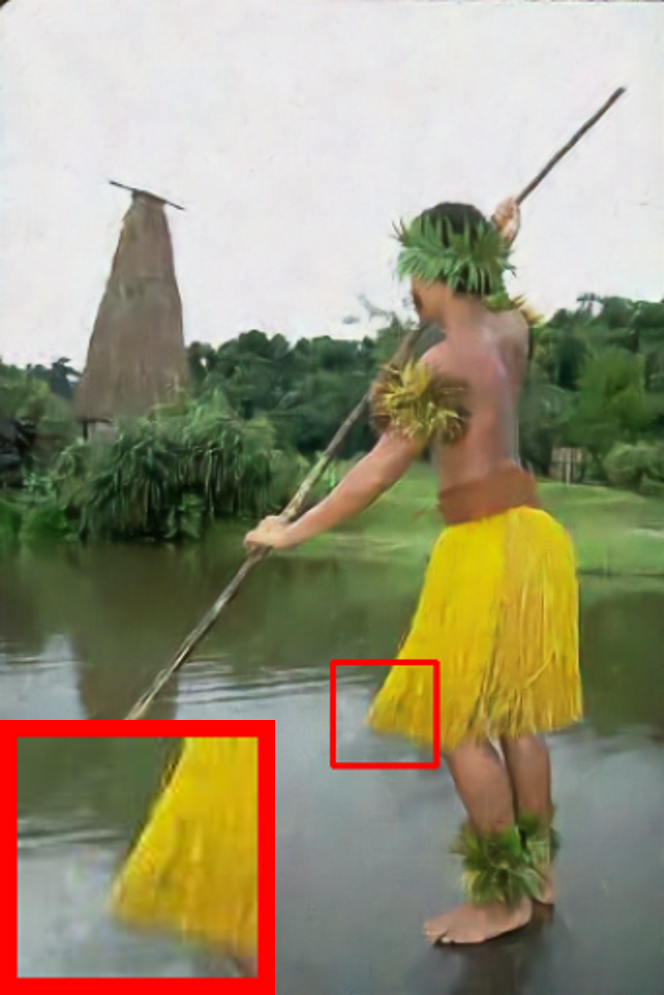}
        \centering  \tiny \textbf{Proposed}\\\textbf{27.21 dB}
        \label{BSDP10}
    \end{minipage}\\
    
\end{tabular}
}
 
\caption{Visual quality comparison on the BSD 68 dataset with Poisson noise with levels $\lambda \in[5; 50]$ (zoom-in for best view).}
 
\label{ BSD Poisson} 
\end{center}
\end{figure}


\begin{figure}[htbp]
\begin{center}
\resizebox{\columnwidth}{!}{%
\begin{tabular}{l l l l l}

    \begin{minipage}{.1\textwidth}
      \includegraphics[width=20mm, height=20mm]{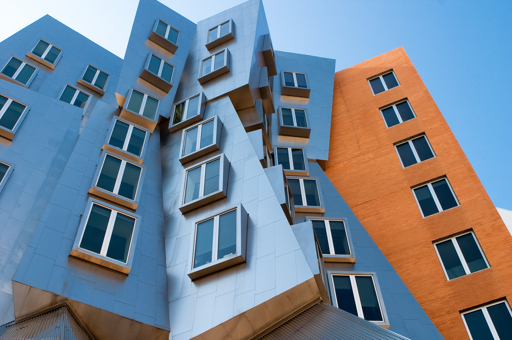}   
         \centering \tiny \textbf{Ground  }\\ \textbf{Truth}
    \label{UR1}
    \end{minipage}
    
    &

    \begin{minipage}{.1\textwidth}
      \includegraphics[width=20mm, height=20mm]{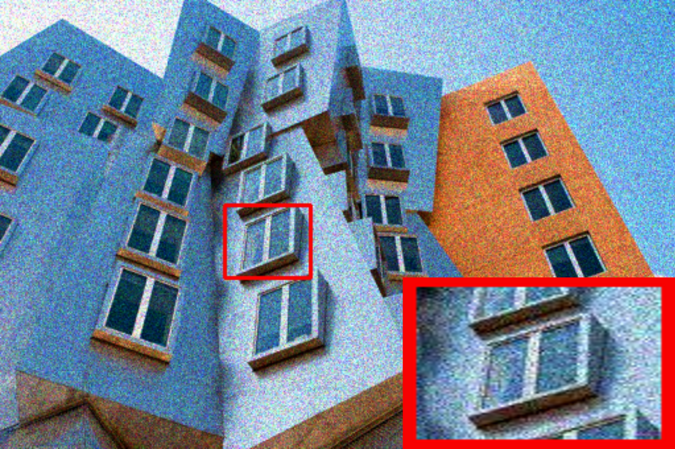}  
         \centering \tiny \textbf{Noisy patch}\\ \textbf{19.27 dB}
    \label{UR2}
    \end{minipage}
    &
    \begin{minipage}{.1\textwidth}
      \includegraphics[width=20mm, height=20mm]{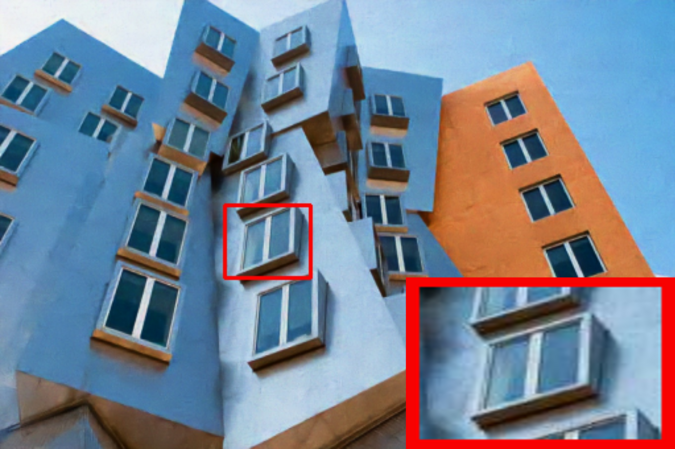}
        \centering \tiny \textbf{DNCNN}\\ \textbf{31.67 dB}
        \label{UR3}
    \end{minipage}
  &
    \begin{minipage}{.1\textwidth}
     \includegraphics[width=20mm, height=20mm]{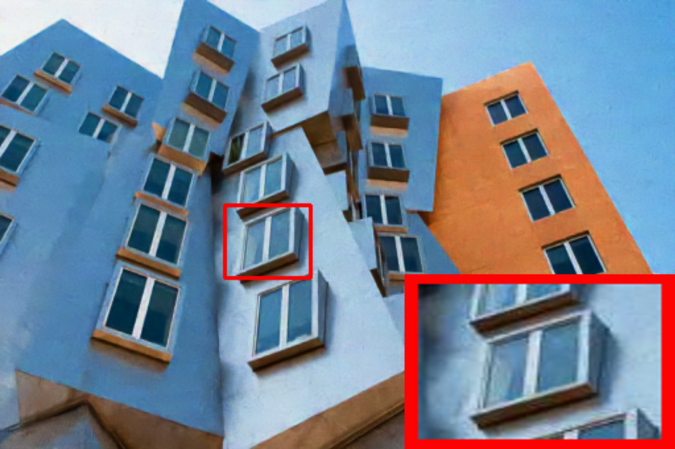}
        \centering  \tiny \textbf{N2N}\\ \textbf{31.20 dB}
        \label{UR4}
    \end{minipage}
    &
    \begin{minipage}{.1\textwidth}
     \includegraphics[width=20mm, height=20mm]{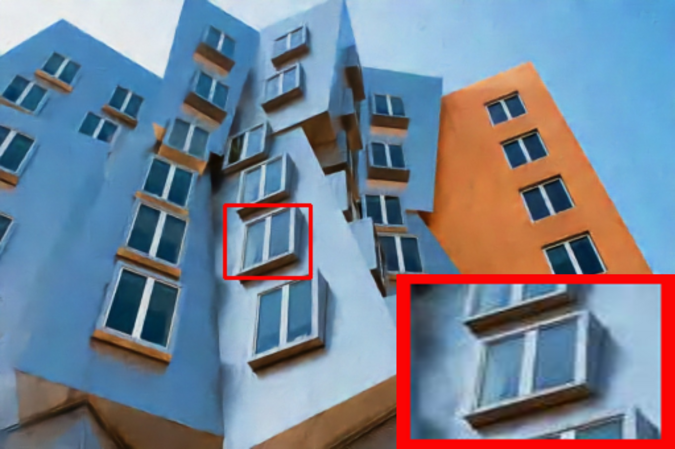}
        \centering  \tiny \textbf{N2S}\\ \textbf{30.22 dB}
        \label{UR5}
    \end{minipage}\\
    \addlinespace
     \begin{minipage}{.1\textwidth}
     \includegraphics[width=20mm, height=20mm]{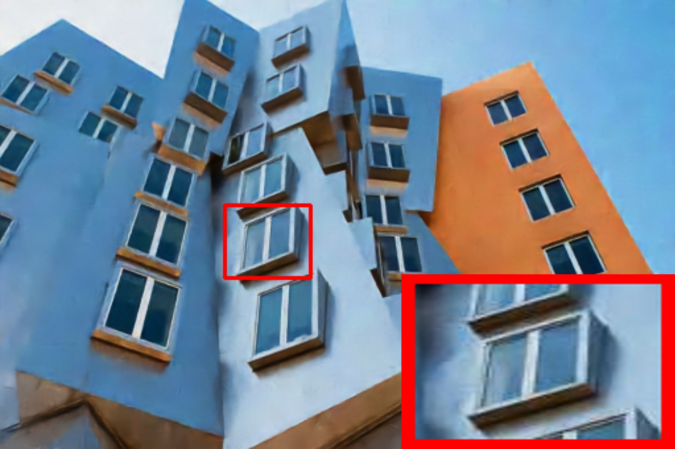}
        \centering  \tiny \textbf{N2V}\\ \textbf{30.50 dB}
        \label{UR6}
    \end{minipage}
    &
     \begin{minipage}{.1\textwidth}
     \includegraphics[width=20mm, height=20mm]{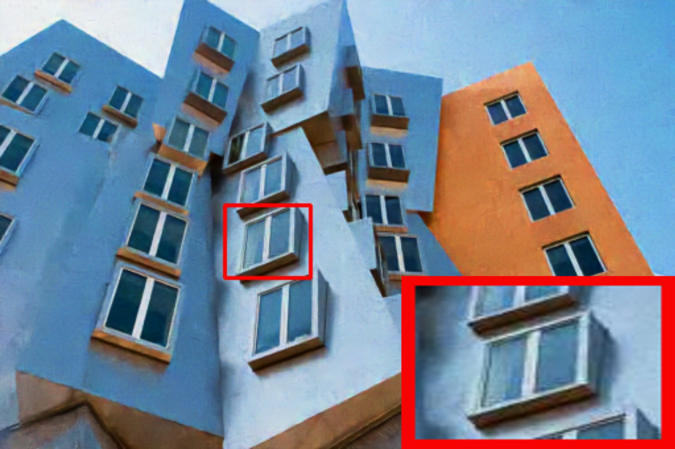}
        \centering  \tiny \textbf{Nr2N}\\ \textbf{30.23 dB}
        \label{UR7}
    \end{minipage}
    &
     \begin{minipage}{.1\textwidth}
     \includegraphics[width=20mm, height=20mm]{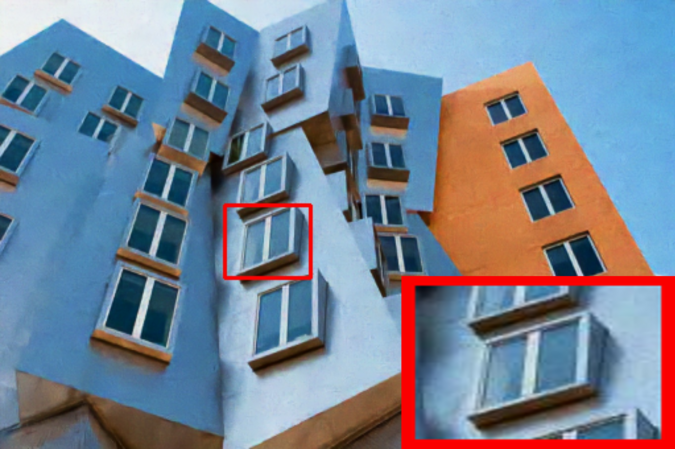}
        \centering  \tiny \textbf{R2R}\\ \textbf{30.79 dB}
        \label{UR8}
    \end{minipage}
    &
     \begin{minipage}{.1\textwidth}
     \includegraphics[width=20mm, height=20mm]{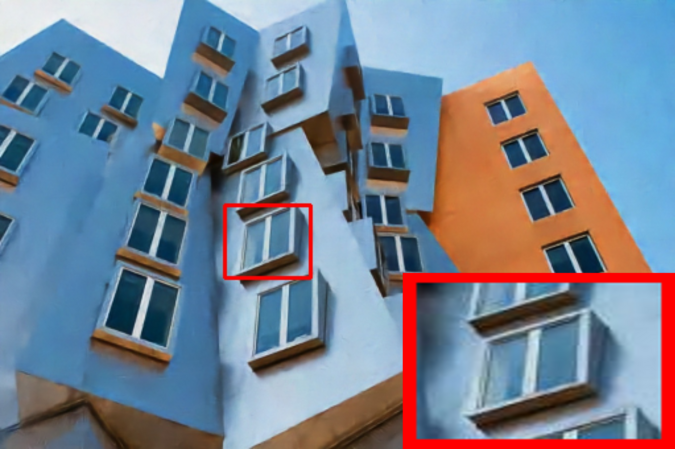}
        \centering  \tiny \textbf{SURE}\\\textbf{30.48 dB} 
        \label{UR9}
    \end{minipage}
    &
     \begin{minipage}{.1\textwidth}
     \includegraphics[width=20mm, height=20mm]{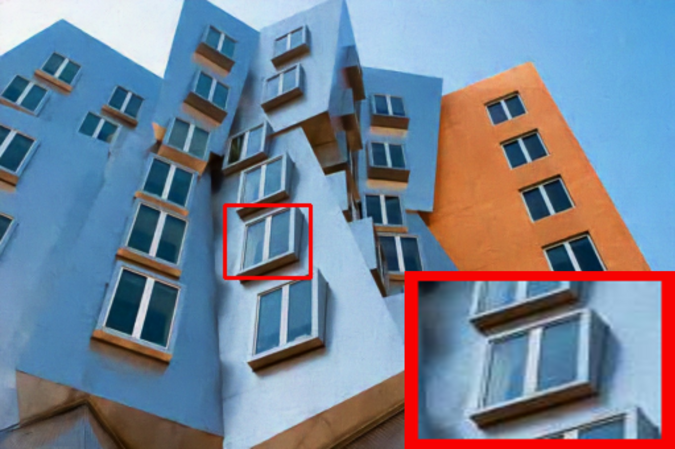}
        \centering  \tiny \textbf{Proposed}\\\textbf{32.56 dB}
        \label{UR10}
    \end{minipage}\\
    
\end{tabular}
}
 
\caption{Visual quality comparison on the UR 100 dataset with Poisson noise with levels $\lambda \in[5; 50]$ (zoom-in for best view).}
 
\label{ UR Poisson} 
\end{center}
\end{figure}


As DIV2K was used as our source of clean images, we applied synthetic noise to generate noisy training images with Gaussian and Poisson noise.  We considered the following three types of synthetic noise distributions: (1) Gaussian noise with a fixed level $\sigma = 25$, (2) Gaussian noise with a fixed level $\sigma = 50$, and, (3) Poisson noise with varied noise levels $\lambda \in[5; 50]$.  Then we utilized  BSD 68, urban100 and Kodak24 image sets as our test sets to perform evaluation.

In the training phase, we have followed separated noise aware training for the both Gaussian and Poisson noise. As mentioned above, we trained with the fixed noise parameter. However, we merged all the fixed noisy observations in a single training, hence a single model is required to obtain clean approximations from the given images. In contrast to other studies, we do not rely on multiple training phases for multiple noise instances. Additionally, we have trained our network with blind setup, where noise parameters varies randomly. In such case, we have achieved similar or higher results from the proposed training setup. Higher results are due to the randomness effect of the blind noise sampling. As most of the studies provided codes and trained network for the fixed noise level, we present the results based upon such availability.

\subsection{Gaussian noise result}

We compared our method against the supervised DnCNN \cite{zhang2017beyond}, Noise2self \cite{batson2019noise2self}, Noise2void \cite{krull2019noise2void}, Noise2Noise \cite{lehtinen2018noise2noise},  Noiser2noise \cite{moran2020noisier2noise}, Recorrupted2recourrpted \cite{pang2021recorrupted}, and SURE \cite{soltanayev2018training}. In the table \ref{Tab:t1}, we have presented our denoising performance for Gaussian noise, with respect to the aforementioned methods. Visual comparisons are present in the figure \ref{ BSD AWGN} and \ref{ KODAK AWGN}. From the PSNR and SSIM metric performance at table \ref{Tab:t1}, proposed approach have obtained better metric performance compared to the supervised baseline, as well as exceeding the contemporary self-supervised alternatives. 

To have a fair discussion about Noise2Noise performance, which usually obtain high PSNR and SSIM score in the original study. Sudden low score is due to the DnCNN network instead of the regular Unet baseline is present in this paper. For supervised baseline, presented score is not the limiting performance of the DnCNN network, further extendable with sufficient data augmentations and optimization tricks. It is surprising that, our result exceeds the supervised baseline for additive Gaussian noise performance. Experimentally we have observed that, proposed pull regularization helps to obtain better performance.

\subsection{Poisson noise result}

 For the Poisson noise, we have followed the similar training strategy as we have done with the Gaussian noise. Only exception made here is with noise level. Instead of a fixed noise rate, we have utilized random sampling from 5 to 50, as the similar studies practiced. Here, we did not exceeds the supervised \cite{zhang2017beyond} baseline, as well as the weakly supervised \cite{lehtinen2018noise2noise} one also. However, we have achieved comparable performance to the weakly supervised \cite{lehtinen2018noise2noise} baseline. On the other hand, our study exceeds the previous self-supervised studies with considerable margin in all of three test cases. Table \ref{Tab:t2} contains our result for the Poisson noise and figure  \ref{ BSD Poisson}  is showing the visual comparison between all of the compared approaches.

\section{Conclusion}
In this paper, we propose a novel self-supervised method for image denoising which doesn't require any noisy-clean pairs, multiple noisy observations, and specific noise modeling. Our approach takes advantage of image degradation through bit compression and row shifting  and allows the CNN to learn the hidden representations of an image by minimizing the loss functions. These adaptive self-supervision loss functions overcome the requirement of zero-mean constraint where prior knowledge of noise statistics is not readily available. Its effectiveness was further demonstrated by the numerical results on AWGN and Poisson noise removal. For AWGN removal our method has obtained competitive results compared to the state-of-the-art supervised learning methods. Extensive experiments showed that the proposed method outperformed existing self-supervised deep denoisers, and is competitive enough as a representative of supervised deep denoisers.

\bibliographystyle{unsrt}  
\bibliography{references}

\begin{thebibliography}{10}

\bibitem{lehtinen2018noise2noise}
Jaakko Lehtinen, Jacob Munkberg, Jon Hasselgren, Samuli Laine, Tero Karras,
  Miika Aittala, and Timo Aila.
\newblock Noise2noise: Learning image restoration without clean data.
\newblock {\em arXiv preprint arXiv:1803.04189}, 2018.

\bibitem{moran2020noisier2noise}
Nick Moran, Dan Schmidt, Yu~Zhong, and Patrick Coady.
\newblock Noisier2noise: Learning to denoise from unpaired noisy data.
\newblock In {\em Proceedings of the IEEE/CVF Conference on Computer Vision and
  Pattern Recognition}, pages 12064--12072, 2020.

\bibitem{batson2019noise2self}
Joshua Batson and Loic Royer.
\newblock Noise2self: Blind denoising by self-supervision.
\newblock In {\em International Conference on Machine Learning}, pages
  524--533. PMLR, 2019.

\bibitem{krull2019noise2void}
Alexander Krull, Tim-Oliver Buchholz, and Florian Jug.
\newblock Noise2void-learning denoising from single noisy images.
\newblock In {\em Proceedings of the IEEE/CVF Conference on Computer Vision and
  Pattern Recognition}, pages 2129--2137, 2019.

\bibitem{xie2020noise2same}
Yaochen Xie, Zhengyang Wang, and Shuiwang Ji.
\newblock Noise2same: Optimizing a self-supervised bound for image denoising.
\newblock {\em Advances in Neural Information Processing Systems},
  33:20320--20330, 2020.

\bibitem{wang2022blind2unblind}
Zejin Wang, Jiazheng Liu, Guoqing Li, and Hua Han.
\newblock Blind2unblind: Self-supervised image denoising with visible blind
  spots.
\newblock In {\em Proceedings of the IEEE/CVF Conference on Computer Vision and
  Pattern Recognition}, pages 2027--2036, 2022.

\bibitem{pang2021recorrupted}
Tongyao Pang, Huan Zheng, Yuhui Quan, and Hui Ji.
\newblock Recorrupted-to-recorrupted: Unsupervised deep learning for image
  denoising.
\newblock In {\em Proceedings of the IEEE/CVF Conference on Computer Vision and
  Pattern Recognition}, pages 2043--2052, 2021.

\bibitem{soltanayev2018training}
Shakarim Soltanayev and Se~Young Chun.
\newblock Training deep learning based denoisers without ground truth data.
\newblock {\em Advances in neural information processing systems}, 31, 2018.

\bibitem{lequyer2021noise2fast}
Jason Lequyer, Reuben Philip, Amit Sharma, and Laurence Pelletier.
\newblock Noise2fast: Fast self-supervised single image blind denoising.
\newblock {\em arXiv preprint arXiv:2108.10209}, 2021.

\bibitem{lee2020noise2kernel}
Kanggeun Lee and Won-Ki Jeong.
\newblock Noise2kernel: Adaptive self-supervised blind denoising using a
  dilated convolutional kernel architecture.
\newblock {\em arXiv preprint arXiv:2012.03623}, 2020.

\bibitem{buades2005non}
Antoni Buades, Bartomeu Coll, and J-M Morel.
\newblock A non-local algorithm for image denoising.
\newblock In {\em 2005 IEEE Computer Society Conference on Computer Vision and
  Pattern Recognition (CVPR'05)}, volume~2, pages 60--65. IEEE, 2005.

\bibitem{dabov2007image}
Kostadin Dabov, Alessandro Foi, Vladimir Katkovnik, and Karen Egiazarian.
\newblock Image denoising by sparse 3-d transform-domain collaborative
  filtering.
\newblock {\em IEEE Transactions on image processing}, 16(8):2080--2095, 2007.

\bibitem{scetbon2021deep}
Meyer Scetbon, Michael Elad, and Peyman Milanfar.
\newblock Deep k-svd denoising.
\newblock {\em IEEE Transactions on Image Processing}, 30:5944--5955, 2021.

\bibitem{zhang2017beyond}
Kai Zhang, Wangmeng Zuo, Yunjin Chen, Deyu Meng, and Lei Zhang.
\newblock Beyond a gaussian denoiser: Residual learning of deep cnn for image
  denoising.
\newblock {\em IEEE transactions on image processing}, 26(7):3142--3155, 2017.

\bibitem{zhang2018ffdnet}
Kai Zhang, Wangmeng Zuo, and Lei Zhang.
\newblock Ffdnet: Toward a fast and flexible solution for cnn-based image
  denoising.
\newblock {\em IEEE Transactions on Image Processing}, 27(9):4608--4622, 2018.

\bibitem{guo2019toward}
Shi Guo, Zifei Yan, Kai Zhang, Wangmeng Zuo, and Lei Zhang.
\newblock Toward convolutional blind denoising of real photographs.
\newblock In {\em Proceedings of the IEEE/CVF conference on computer vision and
  pattern recognition}, pages 1712--1722, 2019.

\bibitem{zhuo2019ridnet}
Shengkai Zhuo, Zhi Jin, Wenbin Zou, and Xia Li.
\newblock Ridnet: recursive information distillation network for color image
  denoising.
\newblock In {\em Proceedings of the IEEE/CVF International Conference on
  Computer Vision Workshops}, pages 0--0, 2019.

\bibitem{ren2021adaptive}
Chao Ren, Xiaohai He, Chuncheng Wang, and Zhibo Zhao.
\newblock Adaptive consistency prior based deep network for image denoising.
\newblock In {\em proceedings of the IEEE/CVF conference on computer vision and
  pattern recognition}, pages 8596--8606, 2021.

\bibitem{elad2006image}
Michael Elad and Michal Aharon.
\newblock Image denoising via sparse and redundant representations over learned
  dictionaries.
\newblock {\em IEEE Transactions on Image processing}, 15(12):3736--3745, 2006.

\bibitem{quan2020self2self}
Yuhui Quan, Mingqin Chen, Tongyao Pang, and Hui Ji.
\newblock Self2self with dropout: Learning self-supervised denoising from
  single image.
\newblock In {\em Proceedings of the IEEE/CVF conference on computer vision and
  pattern recognition}, pages 1890--1898, 2020.

\bibitem{huang2021neighbor2neighbor}
Tao Huang, Songjiang Li, Xu~Jia, Huchuan Lu, and Jianzhuang Liu.
\newblock Neighbor2neighbor: Self-supervised denoising from single noisy
  images.
\newblock In {\em Proceedings of the IEEE/CVF Conference on Computer Vision and
  Pattern Recognition}, pages 14781--14790, 2021.

\bibitem{xu2020noisy}
Jun Xu, Yuan Huang, Ming-Ming Cheng, Li~Liu, Fan Zhu, Zhou Xu, and Ling Shao.
\newblock Noisy-as-clean: Learning self-supervised denoising from corrupted
  image.
\newblock {\em IEEE Transactions on Image Processing}, 29:9316--9329, 2020.

\end{thebibliography}

\end{document}